\documentclass[a4paper,11pt]{article}
\usepackage{jinstpub}
\usepackage{booktabs}
\usepackage{textgreek}
\usepackage{booktabs,threeparttable}
\usepackage{bm}
\newcommand{\baselinefinder}{\textit{Baseline Finder}\ }
\newcommand{\gnnfinder}{\textit{GNN Finder}\ }
\newcommand{\baselinefitter}{\textit{Baseline Fitter}\ }
\newcommand{\gnnfitter}{\textit{GNN Fitter}\ }
  
\title{\boldmath DCTracks: An Open Dataset for Machine Learning-Based Drift Chamber Track Reconstruction}

\author[a,b]{Liyan Qian,}
\author[a,b*]{Yao Zhang,\note{* Corresponding author.}}
\author[a,b**]{Ye Yuan,\note{** Corresponding author.}}
\author[a,b]{Zhaoke Zhang,}
\author[c]{Jin Fang,}
\author[d]{Shimiao Jiang,}
\author[c]{Jin Zhang,}
\author[a,b]{Ke Li,}
\author[a,b]{Beijiang Liu,}
\author[e,b]{Chenglin Xu,}
\author[e,b]{Yifan Zhang,}
\author[f]{Xiaoqian Jia,}
\author[f]{Xiaoshuai Qin}
\author[f]{and Xingtao Huang}

\affiliation[a]{Institute of High Energy Physics, Chinese Academy of Sciences\\No.19B Yuquan Road, Shijingshan, Beijing, China} 
\affiliation[b]{University of Chinese Academy of Sciences\\No.19A Yuquan Road, Shijingshan, Beijing, China}

\affiliation[c]{Sun Yat-sen University\\School of Science, Shenzhen Campus of Sun Yat-sen University, Shenzhen, China}
\affiliation[d]{China Academy of Space Technology\\No.104 Youyi Road, Haidian, Beijing, China}
\affiliation[e]{Institute of Automation Chinese Academy of Sciences\\No.95 Zhongguancun East Road, Haidian, Beijing, China‌}
\affiliation[f]{Key Laboratory of Particle Physics and Particle Irradiation (MOE), Institute of Frontier and Interdisciplinary Science, Shandong University,\\Qingdao, Shandong, China}

\emailAdd{zhangyao@ihep.ac.cn, yuany@ihep.ac.cn}

\abstract{We introduce a Monte Carlo~(\textsc{MC}) dataset of single- and two-track drift chamber events to advance  Machine Learning~(\textsc{ML})-based track reconstruction. To enable standardized and comparable evaluation, we define track reconstruction specific metrics and report results for traditional track reconstruction algorithms and a Graph Neural Networks ~(\textsc{GNNs}) method, facilitating rigorous, reproducible validation for future research.}


\begin{document}
\maketitle
\flushbottom

\section{Introduction}
\label{sec:introduction}
Precision tests of the Standard Model and searches for physics beyond it rely on the high energy physics experiments. To achieve the physics goals of the experiments, high precision detectors and advanced data analysis are both essential. In particular, it relies on the precise charged particle reconstruction through pattern recognition and track fitting. As high energy physics experiments face rising instantaneous luminosity, detector upgrades and increasingly stringent demands on data-simulation statistical compatibility, track reconstruction must maintain accuracy, processing speed and robustness under complex final-state conditions as well as detector imperfections\cite{CGEMupgrade,BelleIIupgrade,BESIIIlumi}. Key challenges in charged particle track reconstruction include background suppression, integrating track reconstruction across subdetectors, improved efficiency for low-momentum and displaced tracks, reducing clone and fake rates, and improving data–simulation agreement.

Traditional track reconstruction relies mainly on mature pattern recognition \cite{pattern} and Kalman filter-based track fitting algorithms \cite{fitting}. Recently, Machine Learning~(\textsc{ML}) approaches --- particularly Graph Neural Networks~(\textsc{GNNs}) offer significant benefits for track reconstruction by enabling end-to-end learning of track parameters such as momentum, direction and associated hits directly from raw detector data. This capability allows for direct optimization of key evaluation metrics, making them a highly promising approach for track reconstruction \cite{Correia:2025deq,Reuter:2024kja,Plini:2025hyu}. However, the shortage of publicly available datasets and specific evaluation metrics remains a major barrier: it impedes reproducible testing and fair comparison across studies, and it largely discourages participation from the broader ML community. In this context, interdisciplinary collaboration and open datasets are essential to realizing the full potential of ML-based track reconstruction.

We address the shortage of public datasets by releasing a drift chamber dataset with full Monte Carlo (MC) and detector response, followed by a preprocessing pipeline (see section "\nameref{sec:Dataset_for_drift_chambers}"). 
To support fair comparison, we also propose specific evaluation metrics (see section "\nameref{sec:Evaluation_metrics}"). 

Subsequently, we implement a ML track reconstruction model based on \textsc{GNNs} \cite{Reuter:2024kja} and compare it with traditional methods (see section "\nameref{sec:Benchmark_experiments}"). 
The results confirm the reliability of this dataset and the effectiveness of the evaluation metrics, establishing a robust, open platform for future research in ML-based track reconstruction.

\section{Related work}
\label{sec:related_work}
ML-based track reconstruction has achieved notable progress \cite{GNN1,GNN2,GNN3} and a handful of public datasets have emerged to support this research. Different research teams use varied datasets and evaluation metrics, hindering direct comparison of model performance. 

The dataset of TrackML Particle Tracking Challenge \cite{trackML}, as utilized by Samuel Van Stroud \cite{TrackML1} and Rusov, D. I. \cite{TrackML2}, is generated from a generalized LHC-like detector and provides its evaluation metrics. Each event simulates one hard top quark-antiquark pair ($t\bar{t}$) interaction overlaid with an additional 200 soft QCD interactions, which reproduces the high pileup conditions expected at the HL-LHC \cite{LHC}. About $10^4$ particles and $10^5$ hits simulated in an event.  
To develop and evaluate particle reconstruction algorithms, Lukas Heinrich et al. \cite{GNN1} utilize the OpenDataDetector (ODD) \cite{ODD1} to generate simulation events. This virtual hermetic detector is designed to serve as a template for (HL-)LHC-style particle detectors, providing a standardized framework for algorithm research and development. In their work, they generate top-antitop quark pair ($t\bar{t}$) events with a pile-up of 200, corresponding to challenging high-multiplicity scenarios typical of collider environments. Recently, the ColliderML dataset \cite{ColliderML} was released as a large-scale, open, experiment-agnostic resource for high-luminosity LHC physics. It provides over one million fully simulation events across ten Standard Model and Beyond Standard Model processes, with realistic pile-up overlay and ODD-based detector geometry. While ColliderML fills critical gaps for HL-LHC-oriented ML research, it still targets the high-pileup, high-multiplicity environment of future hadron colliders, which is fundamentally different from the low-background, low-multiplicity scenarios of precision experiments.

Unlike the high‑energy‑frontier environment of the HL‑LHC, precision flavor factories (e.g., BESIII \cite{BESIII} and BelleII \cite{BelleII}) operate with much lower backgrounds and prioritize precision measurements. They feature substantially lower event multiplicity, cleaner event topologies and stricter requirements on momentum resolution and tracking efficiency, especially for low‑momentum particles. This creates a gap: there is a shortage of datasets with simple track topologies that faithfully capture drift chamber characteristics—an essential resource for the fundamental validation and iterative development of track reconstruction methods tailored to $\tau$-charm experiments. 
We aim to establish such a dataset to accelerate the development of ML methods for track reconstruction in high-precision physics experiments.

\section{The cylindrical multilayer drift chamber}
\label{sec:MDC}

The cylindrical multilayer drift chamber is tasked with measuring the momentum and position of the tracks for final-state charged particles and identifying particle species by measuring the ionization energy loss (dE/dx) of charged particles in the gas. It is widely adopted in high energy physics experiments including BESIII, CEPC~\cite{CEPC}, STCF~\cite{STCF}, BelleII, COMET~\cite{COMET}, MEGII~\cite{MEGII}, FCC~\cite{FCC}. 

Our dataset is based on the Multilayer Drift Chamber (\textsc{MDC}) \cite{MDC} of the BESIII spectrometer. The BESIII at the Beijing Electron Positron Collider \textsc{II} (\textsc{BEPCII})\cite{BEPCII} is located in Beijing, China, conducts particle physics research in the $\tau$-charm energy region. Since 2009, \textsc{BEPCII} has accumulated approximately 10 billion $J/\psi$ events, 2.7 billion $\psi(2S)$ events and 20.3~fb$^{-1}$ of data at the $\psi(3770)$ resonance \cite{BESIIIdata}. 

\begin{figure}[!htbp]
  \centering
  \begin{minipage}{0.38\textwidth}
    \centering
    \includegraphics[width=\linewidth]{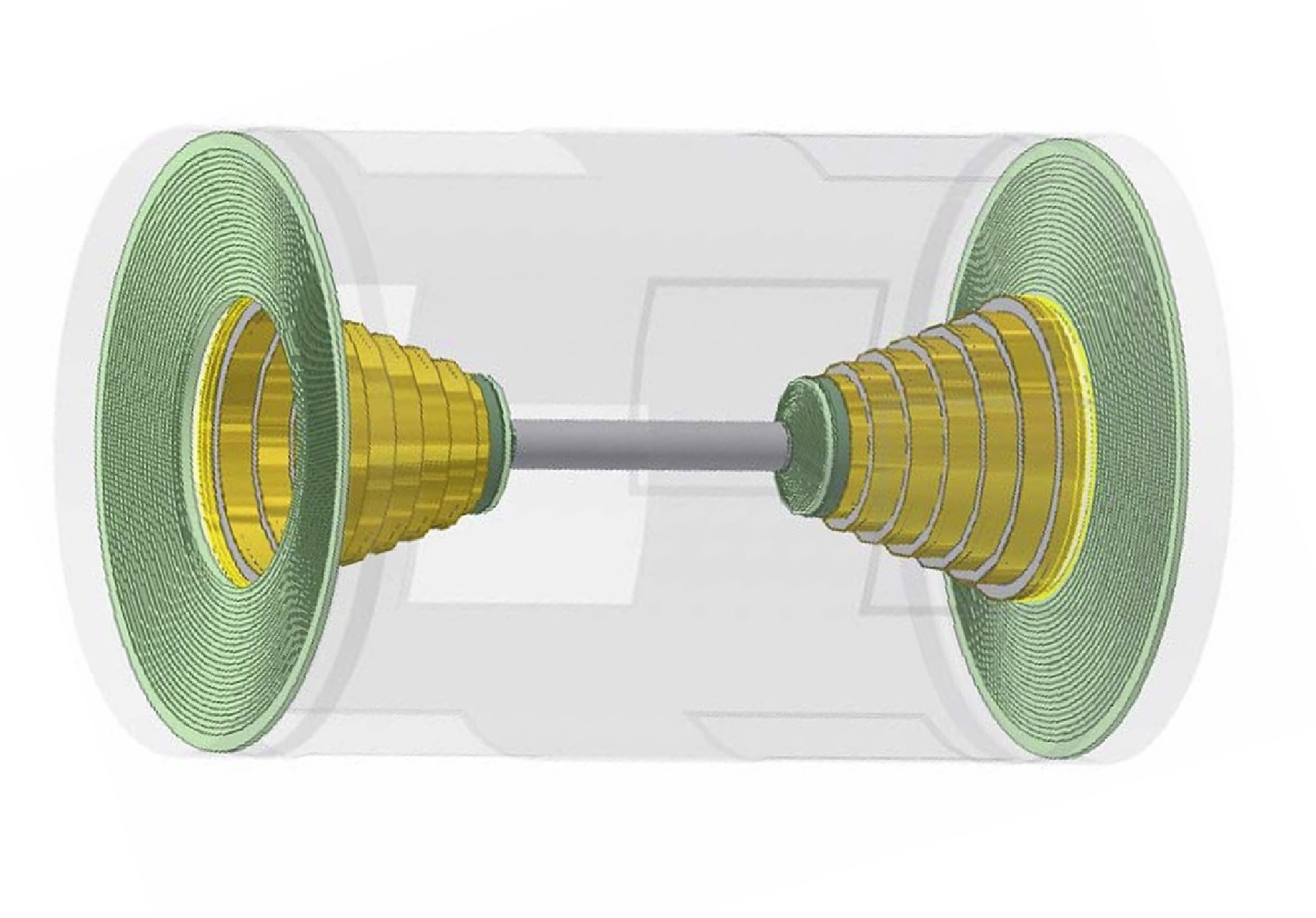}
    % \vspace{1pt}
    \label{fig:3D_display_a}
  \end{minipage}
  \hspace{1cm}
  \begin{minipage}{0.301\textwidth}
    \centering
    \includegraphics[width=\linewidth]{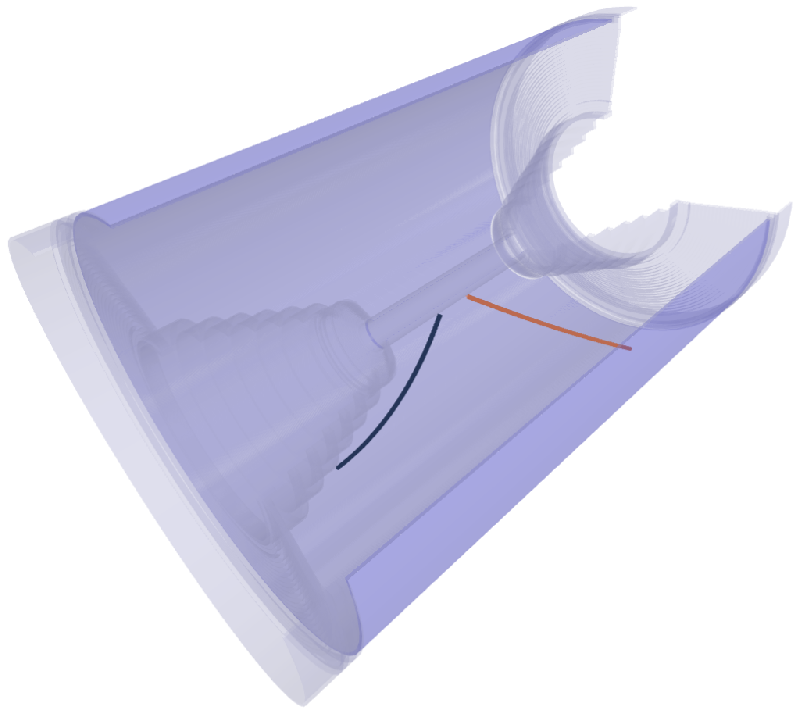}
    % \vspace{1pt}
    \label{fig:3D_display_b}
  \end{minipage}
  \caption{BESIII MDC structure (left) and 3D view of the event (right).}
  \label{fig:3D_display}
\end{figure}

Figure~\ref{fig:3D_display} shows the structure of MDC and the 3D view of an event. Geometrically, the \textsc{MDC} features a length of 2400~mm, an inner radius of 59~mm and an outer radius of 800~mm, with a polar angle coverage in $-0.93 < \cos \theta < 0.93$. It consists of 6796 drift cells, each with a square-like structure. In terms of wire layering, the \textsc{MDC} has 43 sense-wire layers, grouped into superlayers of four sense-wire layers each, except for the outermost superlayer, which contains 3 layers (see table~\ref{tab:MDC_layer}). The MDC operates in a ~$1.0$\,T magnetic field with a helium-based gas mixture as the working medium. The design single-wire spatial resolution is about $130\,\mu$m and a transverse momentum resolution = 0.5\% at 1 GeV/$c$. 

\begin{table}[htbp]
    \centering
    \small
    \setlength{\tabcolsep}{4pt}
    \begin{threeparttable}
    \caption{MDC layer structure and geometry parameters.}
    \label{tab:MDC_layer}
    \begin{tabular}{c c c c c c }
        \toprule
        \textbf{Superlayer} & 
        \textbf{Type} & 
        \bm{$N_{\mathrm{layer}}$}&
        \bm{$N_{\mathrm{wire}}/\mathrm{layer}$} &
        \textbf{Radius (mm)} &        
        \textbf{Length (mm)}  \\
        \midrule
        1  & U & 4 & 40,44,48,56     & $\sim$79 -- 115  & 780--816 \\
        2  & V & 4 & 64,72,80,80     & $\sim$127 -- 162 & 828 -- 864 \\
        3  & A & 4 & 76,76,88,88     & $\sim$197 -- 246 & 1092 --1272  \\
        4  & A & 4 & 100,100,112,112 & $\sim$262 -- 311 & 1442 -- 1612 \\
        5  & A & 4 & 128,128,140,140 & $\sim$327 -- 375 & 1782 -- 1952 \\
        6  & U & 4 & 160$\times$4 & $\sim$400 -- 448 & 2174 -- 2192 \\
        7  & V & 4 & 176$\times$4 & $\sim$464 -- 514 & 2198 -- 2216 \\
        8  & U & 4 & 208$\times$4 & $\sim$530 -- 579 & 2222 -- 2240 \\
        9  & V & 4 & 240$\times$4 & $\sim$595 -- 642 & 2246 -- 2264 \\
        10 & A & 4 & 256$\times$4 & $\sim$667 -- 716 & 2276 -- 2294 \\
        11 & A & 3 & 288$\times$3 & $\sim$732 -- 763 & 2300 -- 2306\\
        \bottomrule
    \end{tabular}
    \begin{tablenotes}[flushleft]
    \footnotesize
    \item Notation: a$\times$n denotes n layers each with number of wire a. A: axial superlayers, U: stereo superlayers with negative tilt angle, V: Stereo superlayers with positive tilt angle.
    \end{tablenotes}
\end{threeparttable}
\end{table}

\newpage
\section{Dataset for drift chambers}

\label{sec:Dataset_for_drift_chambers}
\subsection{Event simulation}
The dataset in this work is generated using a GEANT4-based full simulation \cite{Simulation} in the BESIII Offline Software System (BOSS) \cite{BOSS}. To support foundational research and reduce complexity of track reconstruction algorithm, \textit{single-track} and \textit{two-track} events are included.
To avoid complications from curled tracks in the MDC, we impose a requirement of transverse momentum $p_\mathrm{T}$ > 0.15 GeV. The detailed simulation settings are listed in table~\ref{table:dataset_params}. We plan to include dedicated support for low‑$p_{\mathrm{T}}$ curled tracks in future updates.

\begin{table}[!htbp]
    \centering
    % 核心：设置caption到表格的间距为6pt（和MDC表格默认值一致）
    \setlength{\belowcaptionskip}{6pt}  
    % 保持caption上方间距为默认值（和MDC表格一致）
    \setlength{\abovecaptionskip}{10pt}  
    \caption{Kinematic settings for \textit{single-track} and \textit{two-track} event simulation.}
    \label{table:dataset_params} % caption后紧跟label，符合LaTeX最佳实践
    \small % 匹配jinstpub的表格字体大小
    \begin{tabular}{@{}c c c c c@{}} 
        \toprule[1.2pt]
        Event Type              & $p_T$ [GeV/$c$] & $\cos\theta$ & $\phi$ [rad] & Particles \\
        \midrule[0.8pt]
        Single-track            & $0.15\sim1.5$   & $-0.93\sim0.93$ & $0\sim2\pi$         & $e^\pm, \mu^\pm, \pi^\pm, K^\pm, p, \bar{p}$ \\
        Conventional two-track  & $0.15\sim1.5$   & $-0.93\sim0.93$ & $0\sim2\pi$         & $\pi^+\pi^-$ \\
        Close-by two-track      & $0.15\sim1.5$   & $-0.93\sim0.93$ & $\Delta\phi=0.2$    & $\pi^+\pi^-$ \\
        \bottomrule[1.2pt]
    \end{tabular}
\end{table}

\begin{figure}[!htbp]
    \centering
    % 三张图横向并列，各占0.3\textwidth，与原宽度一致
    \begin{minipage}[t]{0.31\textwidth}
        \centering
        \includegraphics[width=\linewidth]{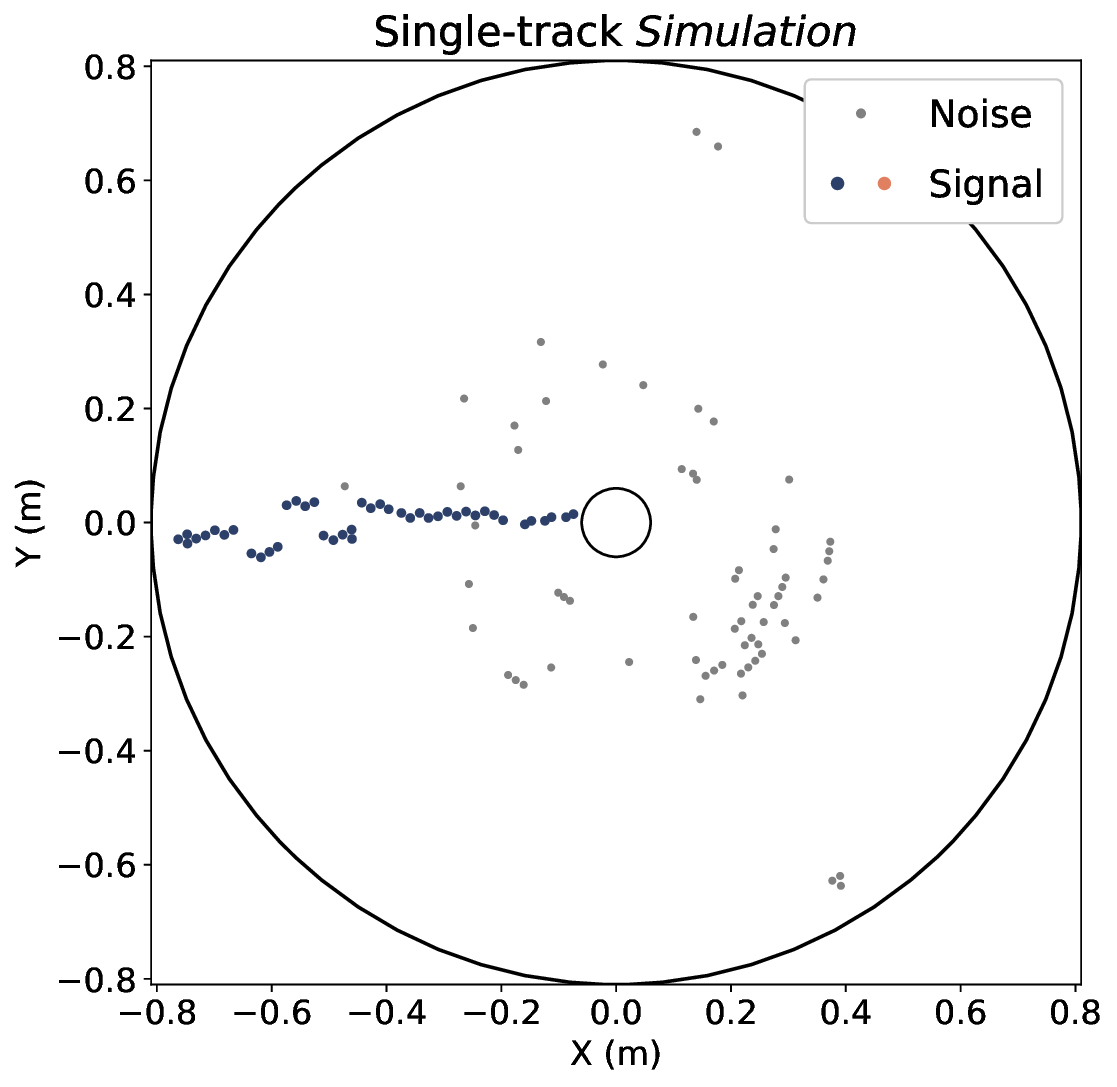}
    \end{minipage}
    \hfill
    \begin{minipage}[t]{0.31\textwidth}
        \centering
        \includegraphics[width=\linewidth]{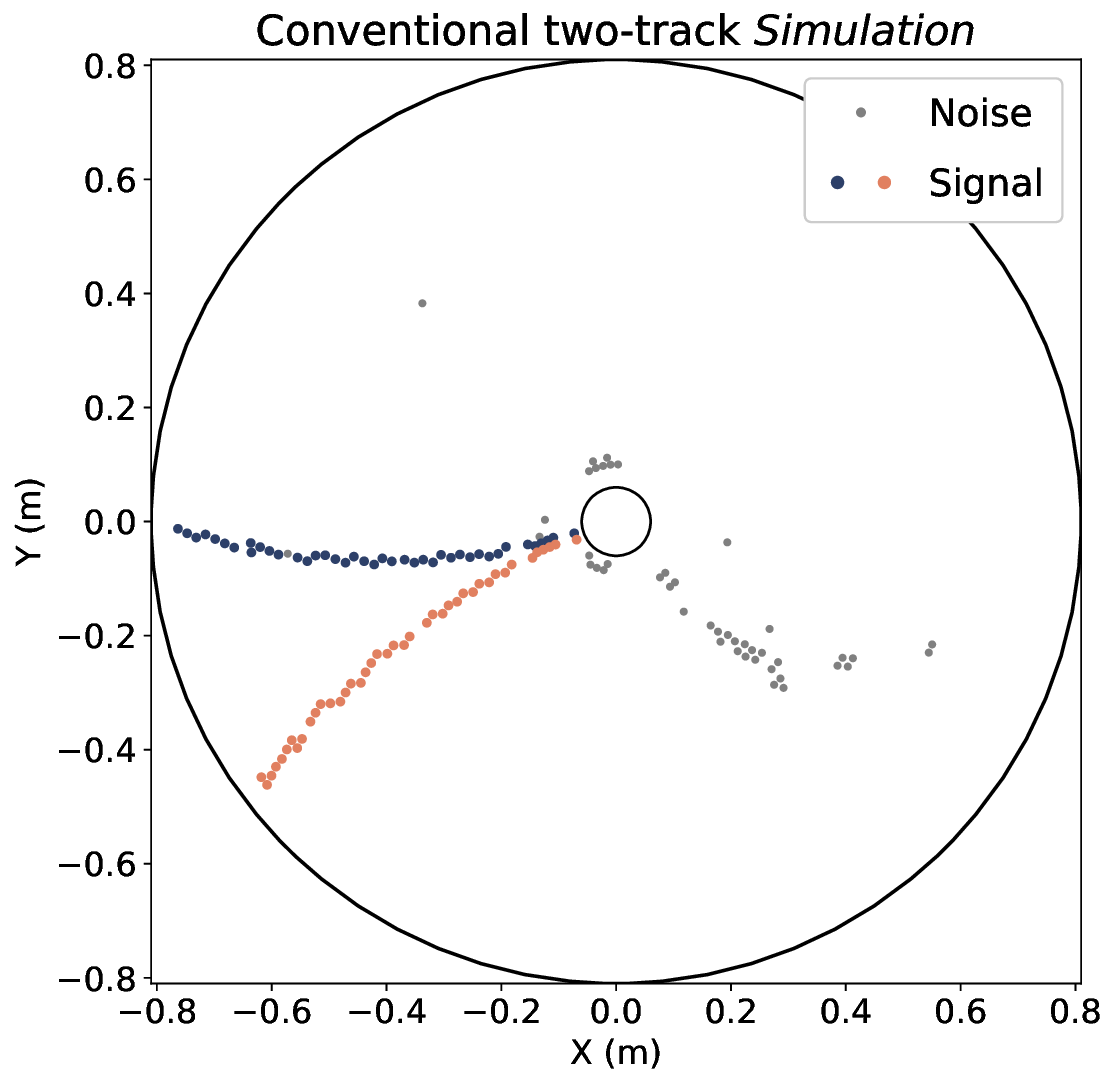}
    \end{minipage}
    \hfill
    \begin{minipage}[t]{0.31\textwidth}
        \centering
        \includegraphics[width=\linewidth]{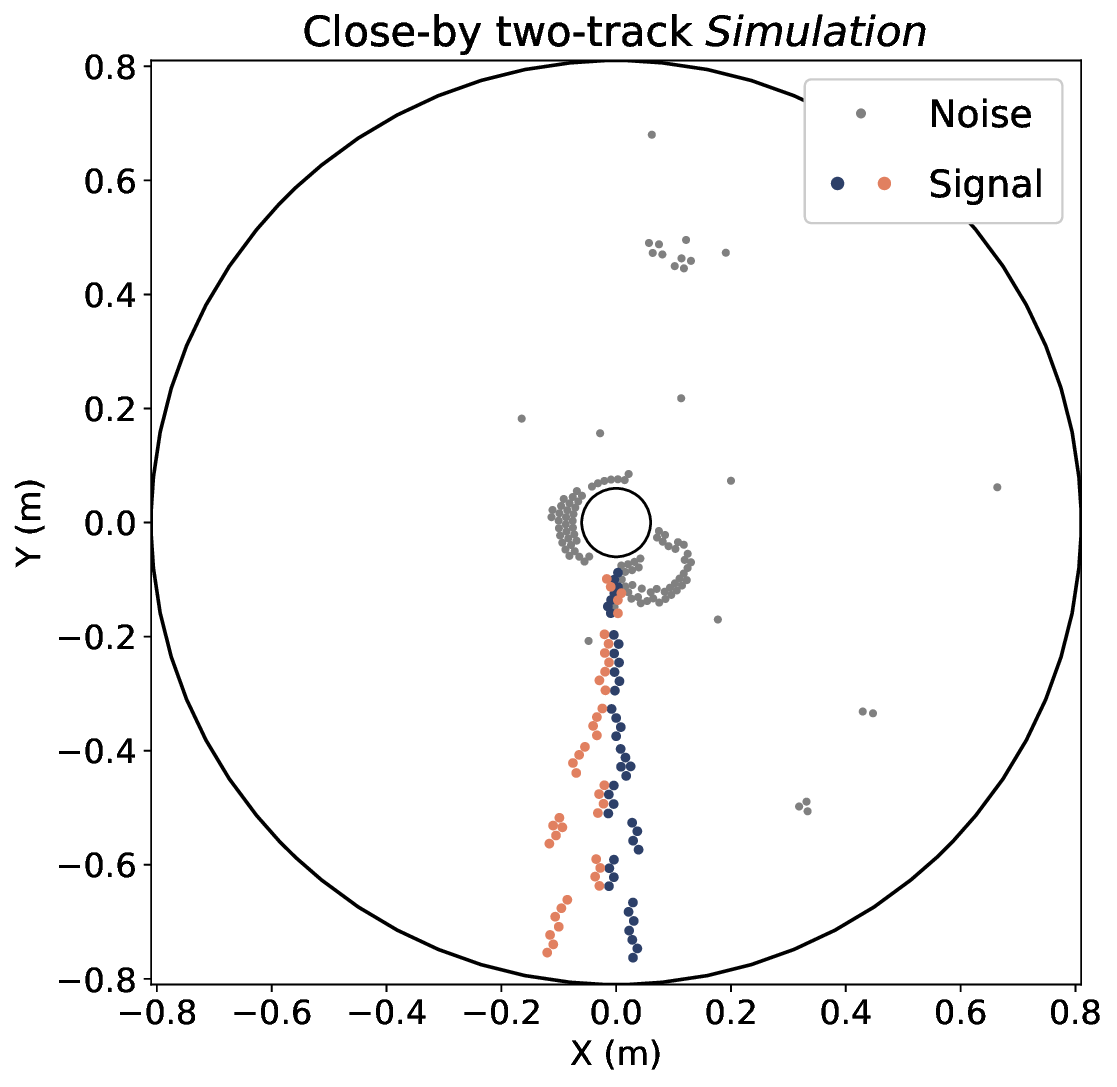}
    \end{minipage}
    
    \caption{Displays of the simulated events in the x-y plane for a \textit{single-track} event (left), a \textit{conventional two-track} event (middle) and a \textit{close-by two-track} event (right).}
    \label{fig:raw_event_display}
\end{figure}

For \textit{single-track} events, each event contains one charged track, as illustrated in figure~\ref{fig:raw_event_display}(a). 
These events include five charged particles species: $e^\pm$, $\mu^\pm$, $\pi^\pm$, $K^\pm$, $p$ and $\bar{p}$. 
To ensure comprehensive and effective model training, all events are generated with kinematic parameters ($p_\mathrm{T}$, $\cos\theta$, $\phi$) sampled uniformly over the accessible phase space.

For \textit{two-track} events, each event contains two charged tracks, as shown in figure~\ref{fig:raw_event_display}(b) and (c). 
These are further categorized into two types: 
\textit{conventional two-track} events in figure~\ref{fig:raw_event_display}(b), where the azimuthal angle difference $\Delta\phi$ between the two tracks is unconstrained, 
and \textit{close-by two-track} events in figure~\ref{fig:raw_event_display}(c), where $\Delta\phi$ between the two tracks is constrained to a narrow range. 
For both types, the kinematic parameters of each individual track in the two-track event are sampled uniformly over the accessible phase space.

To reproduce the experimental conditions, all simulated events are mixed with noise including beam-induced backgrounds and detector noise measured in real data. 

%raw event display

% Here, $\theta$ denotes the polar angle, $\cos\theta$ is the angle between the particle's momentum direction and the z axis; $\phi$ refers to the azimuthal angle, the angle of the particle's momentum direction in the transverse plane.

\subsection{Data preprocessing}
\label{sec:Data_preprocessing}
To build a high-quality dataset, we apply a series of selections to the simulated events. The detailed steps are described as follows:

\textbf{Event-level selection.} Using MC truth, we identify and remove events originating from non-signal processes (e.g., non-target decays). This truth-level veto suppresses background contamination, improves data sample purity and prevents the model from learning spurious correlations unrelated to the signal.

\textbf{Track-level selection. }To ensure that the tracks used for ML training have sufficient number of hits for the fitting (see section "\nameref{sec:track_fitting}"), only those tracks that traverse at least 6 layers in the MDC are retained. For any track that fails to meet this minimum layer requirement, all its corresponding hits are labeled as noise hits. This operation effectively filters out short tracks.

\subsection{Dataset description}
The dataset is stored in Comma-Separated Values (CSV) format, chosen for its excellent compatibility, readability and ease of parsing across various programming languages and analysis frameworks. Each row represents a single detector hit in the MDC. To support noise hit filtering, track finding and global track fitting tasks, this hit-centric dataset has its features and labels defined as follows:

\textbf{Features.} 
Input features are derived from individual drift chamber hit measurements, capturing both the spatial and physical measurement properties of each hit. \textbf{Spatial features} describe the geometric position and hierarchical structure of the sense wire associated with each hit. 

\begin{itemize}
    \item \textbf{\texttt{middleX}, \texttt{middleY}} \\
          The Cartesian coordinates of the mid-point of the sense wire at the two ends of the MDC (in cm).

    \item \textbf{\texttt{layer,slayer,locallayer}} \\
          As described in section "\nameref{sec:MDC}", \texttt{layer} is the global layer index (ranging from 0 to 42, corresponding to the 43 total sense-wire layers); \texttt{slayer} is the superlayer index (ranging from 0 to 10, for the 11 total superlayers); \texttt{locallayer} is the local layer index of its parent superlayer (ranging from 0 to 3 for most superlayers and 0 to 2 for the outermost superlayer).  
\end{itemize}

\textbf{Measurement features} characterize the physical signal recorded from the hit, specifically related to the drift time measurement.
\begin{itemize}
    \item \textbf{\texttt{rawDriftDist,rawDriftDistErr}} \\
          \texttt{rawDriftDist} is the drift distance in the cell, derived from the measured drift time by an initial T-X (time–distance) calibration (in cm); \texttt{rawDriftDistErr} is the estimated uncertainty of \texttt{rawDriftDist} (in cm).
\end{itemize}

\textbf{Labels.} Labels are divided into two levels: hit-level and track-level, allowing the model to first distinguish signal hits from noise at the hit level and learn track-level parameters (e.g., momentum, position and charge).
\textbf{Hit-level labels} are assigned to each individual hit and are primarily used for hit classification, noise suppression and hit-to-track grouping.

\begin{itemize}
    \item \textbf{\texttt{isSignal}} \\
          According to MC truth, \texttt{isSignal} is 1 for signal hits and 0 for noise hits. This label is crucial for training the model to reject noise.

    \item \textbf{\texttt{trackIndex}}  \\
          The unique identifier of the simulated particle to which this hit belongs. Signal hits from the same simulated particle share the same \texttt{trackIndex}, with \texttt{trackIndex} > 0. This label enables supervised learning of hit-to-track association and serves as ground truth for ML-based track reconstruction methods.

    \item \textbf{\texttt{scaledFltLen}}  \\
          The path length along the track from the particle’s production vertex to the hit position, normalized by the circumference of the corresponding helix turn.

    \item \textbf{\texttt{lrAmbig}}  \\
          The hit left-right flag is a binary label indicating on which side of the sense wire the hit lies in the local wire coordinate system.
\end{itemize}

\textbf{Track-level labels} are assigned at the level of each MC simulated particle and provide the track parameters and spatial information needed for the supervised learning.

\begin{itemize}
    \item \textbf{\texttt{initialMomX}, \texttt{initialMomY}, \texttt{initialMomZ}} \\
          The momentum vector components at the point of closest approach (POCA) to the origin $O(0,\,0,\,0)$ (in GeV/$c$)  of the particle. These values serve as ground-truth targets for momentum regression.

    \item \textbf{\texttt{initialPosX}, \texttt{initialPosY}, \texttt{initialPosZ}} \\
          The Cartesian coordinates at the POCA to the origin of the particle (in cm). These values provide ground truth for vertex regression.

    \item \textbf{\texttt{charge}} \\
          Signed charge of the track ($+1$ or $-1$).
\end{itemize}

The separation of hit-level and track-level labels supports various tasks learning, such as binary classification (signal vs.\ noise), clustering (hit clustering for track finding) and regression (track parameters). 

Additional feature and label details are not elaborated on here. Please refer to the official documentation of the dataset, where access instructions are specified in the section "\nameref{sec:dataset_access}".

\subsection{Dataset access}
\label{sec:dataset_access}
For members of the BESIII Collaboration, the dataset is available for direct download from the IHEP AI Platform \cite{HEPAI} at \url{https://ai.ihep.ac.cn}. To support cross-disciplinary collaboration on this dataset,  external researchers may request access by emailing [hepai@ihep.ac.cn] and providing a short description of the research objectives and intended use. Requests are subject to approval by the BESIII Software Group.

\section{Evaluation metrics}
\label{sec:Evaluation_metrics}
To assess track reconstruction performance and facilitate fair comparison among ML-based methods, we introduce a set of specific evaluation metrics. The algorithms for these metrics are available on GitHub at \url{https://github.com/lyqian1220/DCTracksMetrics.git}.

\textbf{Hit efficiency} ($\epsilon_{\text{hit}}$) is defined as the fraction of a particle’s detectable truth hits that are correctly reconstructed and matched to that particle:
\begin{equation}
\epsilon_{\text{hit}} = \frac{N_{\text{hit}}^\text{matched}}{N_{\text{hit}}^{\text{detectable}}}.
\label{eq:hit_efficiency}
\end{equation}

Here, $N_{\text{hit}}^\text{matched}$ is the number of reconstructed hits correctly matched to the particle and $N_{\text{hit}}^{\text{detectable}}$ is the number of readout-eligible MC truth hits from that particle (i.e., after overlay, digitization, thresholding and detector inefficiency losses).

\textbf{Hit purity} ($p_{\text{hit}}$) is defined as the fraction of reconstructed hits assigned to a track that are correctly matched to the originating particle:
\begin{equation}
p_{\text{hit}} = \frac{{N_{\text{hit}}^\text{matched}}}{N_{\text{hit}}^{\text{assigned}}}.
\label{eq:hit_purity}
\end{equation}

Here, $N_{\text{hit}}^{\text{assigned}}$ is the total number of reconstructed hits assigned to the track.

\textbf{Track efficiency} ($\epsilon_{\text{track}}$) is defined as the fraction of detectable truth tracks for which a matched reconstructed track exists:
\begin{equation}
\epsilon_{\text{track}} = \frac{N_{\text{track}}^\text{matched}}{N_{\text{track}}^{\text{detectable}}}.
\label{eq:track_efficiency}
\end{equation}

Here, we define a simulated particle a detectable truth track if it has at least six detectable truth hits. $N_{\mathrm{track}}^{\mathrm{detectable}}$ denotes the number of detectable truth tracks in the samples and $N_{\mathrm{track}}^{\mathrm{matched}}$ denotes the subset that have a matched reconstructed track.

A reconstructed track is considered to be a \textit{matched track} if it satisfies the track-matching criteria: $p_{\text{hit}} > 0.50$, $\epsilon_{\text{hit}} > 0.20$ and $N_{\text{hit}}^{\text{matched}} \geq 6$. The $p_{\text{hit}}$ threshold enforces hit purity—at least half of the hits assigned to the reconstructed track must originate from the same truth track; the $\epsilon_{\text{hit}}$ threshold enforces hit efficiency—a minimum fraction of that truth track’s detectable truth hits must be recovered; and the $N_{\text{hit}}^\text{matched}$ requirement ensures a minimum number of hits for a stable helix fit and suppresses spurious candidates such as hit‑sharing artifacts and random combinations. We define a \textit{fake track} as one that fails to satisfy the requirements for purity or efficiency.
 If multiple reconstructed tracks satisfy the matching criteria for the same detectable truth track, the candidate with the highest $\epsilon_{\text{hit}}$ is retained as the matched one and the remainder are termed \textit{clone tracks}.

\textbf{Track charge efficiency} ($\epsilon_{\text{track,q}}$)  is defined as the fraction of detectable truth tracks that are reconstructed with the correct charge:
\begin{equation}
\epsilon_{\text{track,q}} = \frac{N_{\text{track}}^\text{matched,q-correct}}{N_{\text{track}}^{\text{detectable}}}.
\label{eq:track_charge_efficiency}
\end{equation}

\textbf{Wrong charge rate} ($R_{\text{wrong,q}}$) is defined as the fraction of detectable truth tracks that are reconstructed with the wrong charge:
\begin{equation}
R_{\text{wrong,q}} = \frac{N_{\text{track}}^\text{matched,q-incorrect}}{N_{\text{track}}^{\text{detectable}}}.
\label{eq:wrong_charge_rate}
\end{equation}

\textbf{Clone rate} ($R_{\text{clone}}$) is defined as the total count of clone tracks divided by the total count of detectable truth tracks:
\begin{equation}
R_{\text{clone}} = \frac{N_{\text{track}}^\text{clone}}{N_{\text{track}}^{\text{detectable}}}.
\label{eq:clone_rate}
\end{equation}
Here, $N_{\mathrm{track}}^{\mathrm{clone}} $ denotes the total number of clone  tracks in the samples.

\textbf{Fake rate} ($R_{\text{fake}}$) is defined as the total number of fake tracks divided by the total number of detectable truth tracks:
\begin{equation}
R_{\text{fake}} = \frac{N_{\text{track}}^\text{fake}}{N_{\text{track}}^{\text{detectable}}}.
\label{eq:fake_rate}
\end{equation}
Here, $N_{\mathrm{track}}^{\mathrm{fake}} $ denotes the total number of fake tracks in the samples.

To characterize the performance of the track finding and track fitting stages of track reconstruction separately, we define two metric sets. For track finding, we report the \textit{track finding efficiency}, \textit{track charge finding efficiency}, \textit{clone finding rate}, \textit{fake finding rate} and \textit{wrong charge finding rate}. For track fitting, we report the \textit{track fitting efficiency}, \textit{track charge fitting efficiency}, \textit{clone fitting rate}, \textit{fake fitting rate} and \textit{wrong charge fitting rate}.

Finally, we evaluate the precision of the matched reconstructed track parameters, focusing on the transverse momentum $p_{\text{T}}$ for tracks with correct charge.
The normalized residual is defined as
\begin{equation}
\eta_{p_{\text{T}}} = \frac{p_{\text{T}}^{\text{reco}} - p_{\text{T}}^{\text{MC}}}{p_{\text{T}}^{\text{MC}}}.
\label{eq:pt_residual}
\end{equation}

This quantity $\eta_{p_{\text{T}}}$ represents the relative deviation of the reconstructed $p_{\text{T}}$ from its MC truth value, normalized to the MC truth $p_{\text{T}}$. 

The distribution of $\eta_{p_{\text{T}}}$ is typically Gaussian for unbiased reconstruction. The $p_{\text{T}}$ resolution is then quantified as the 68\% coverage of the absolute residual distribution around its median:
\begin{equation}
r(p_{\text{T}}) = P_{68\%} \Bigl( \bigl| \eta_{p_{\text{T}}} - P_{50\%}(\eta_{p_{\text{T}}}) \bigr| \Bigr),
\label{eq:pt_resolution}
\end{equation}
where $P_q$ denotes the $q$-th quantile of the distribution and $P_{50\%}$ is the median~\cite{Reuter:2024kja}. 
For a normal distribution, this corresponds to the standard deviation.

\section{Benchmark experiments}
\label{sec:Benchmark_experiments}
To validate the effectiveness of our dataset and evaluation metrics, and to establish a unified benchmark for downstream ML-based methods, we conduct a comparative study using two track finding approaches—a traditional method and a ML-based method. Both approaches are evaluated with and without subsequent track fitting. Notably, the results for the ML-based method are preliminary, serving as an exploratory start for future development.

\subsection{Track finding and fitting}
\textbf{Baseline track finding.} 
The baseline track finding (called \baselinefinder in the following) employs traditional track reconstruction algorithms \cite{TSF,curlfinder,hough} in the BOSS to reconstruct track candidates from detector hits, assuming a uniform 1\ T magnetic field and neglecting energy loss and multiple scattering. The \baselinefinder employs pattern dictionary matching, local track segment finding, Hough transform and other techniques.

\textbf{GNN track finding. }The GNN-based track finding method (called \gnnfinder hereafter) adopted in this work follows the end-to-end multi-track reconstruction framework \cite{Reuter:2024kja} proposed by L. Reuter et al. This method processes raw detector hits without prior filtering, simultaneously predicting both the number of track candidates in an event and their track parameters. In a subsequent clustering step, hits are assigned to each predicted track candidate and passed to the track fitting stage.

In our work, for \textit{single-track} events, the five particle species were processed by combining their positively and negatively charged counterparts for training and validation, while the positively and negatively charged particles of each species were tested separately. For the \textit{conventional two-track} and \textit{close-by two-track} events, each category was trained, validated and tested independently.
In terms of dataset scale, all training and validation sets comprise approximately 100,000 simulated events, which were split into training and validation subsets at a ratio of 9:1. For the independent test sets, each kind of \textit{single-track} events (for positive and negative charges respectively) contains about 55,000 events and each two-track subcategory (\textit{conventional} and \textit{close-by}) includes around 25,000 events. In the future, we plan to provide a mixed-event dataset. Researchers can then use this dataset to jointly train, validate and test the model, which may help improve its performance and generalization ability.

\textbf{Track fitting.} 
\label{sec:track_fitting}
The track finder provides initial estimates of track parameters and the associated hits to the subsequent track fitting. First, to improve the quality of the track candidates from the track finder, a Runge-Kutta \cite{RK} fitting corrects the tracks considering the energy loss, the multiple scattering and the non-uniform magnetic field effect. Then, tracks are fitted by \texttt{GenFit} \cite{genfit1, genfit2}, where mass hypotheses are applied by Kalman filter and track parameters are defined at the POCA to the origin. In the following, \baselinefitter refers to the track collection obtained by fitting the outputs of the \textit{Baseline Finder} and likewise for the \textit{GNN Fitter}.

\subsection{Results}
\label{sec:Results}

This section shows a comparison between the \gnnfinder and the \textit{Baseline Finder}, both with and without track fitting. We evaluate the track reconstruction performance over three event categories in our dataset (see table~\ref{table:dataset_params}): \textit{single-track} events, \textit{conventional two-track} events and \textit{close-by two-track} events.
Results for \textit{single-track} events are illustrated using $\pi^+$ as a representative; those for other single-track particle species are found to be similar and are thus omitted for brevity.

The results cover "\nameref{sec:hit_efficiency_purity}" and "\nameref{sec:track_efficiency}", followed by "\nameref{sec:resolution}". 
Hereafter, we use $p_{\mathrm{T}}^{\text{MC}}$ and $\cos\theta^{\text{MC}}$ to denote the transverse momentum and the cosine of the polar angle of detectable truth tracks, respectively.

\begin{figure}[!htbp]
    \centering
    % 三张图横向并列，各占0.3\textwidth，与原宽度一致
    \begin{minipage}[t]{0.31\textwidth}
        \centering
        \includegraphics[width=\linewidth]{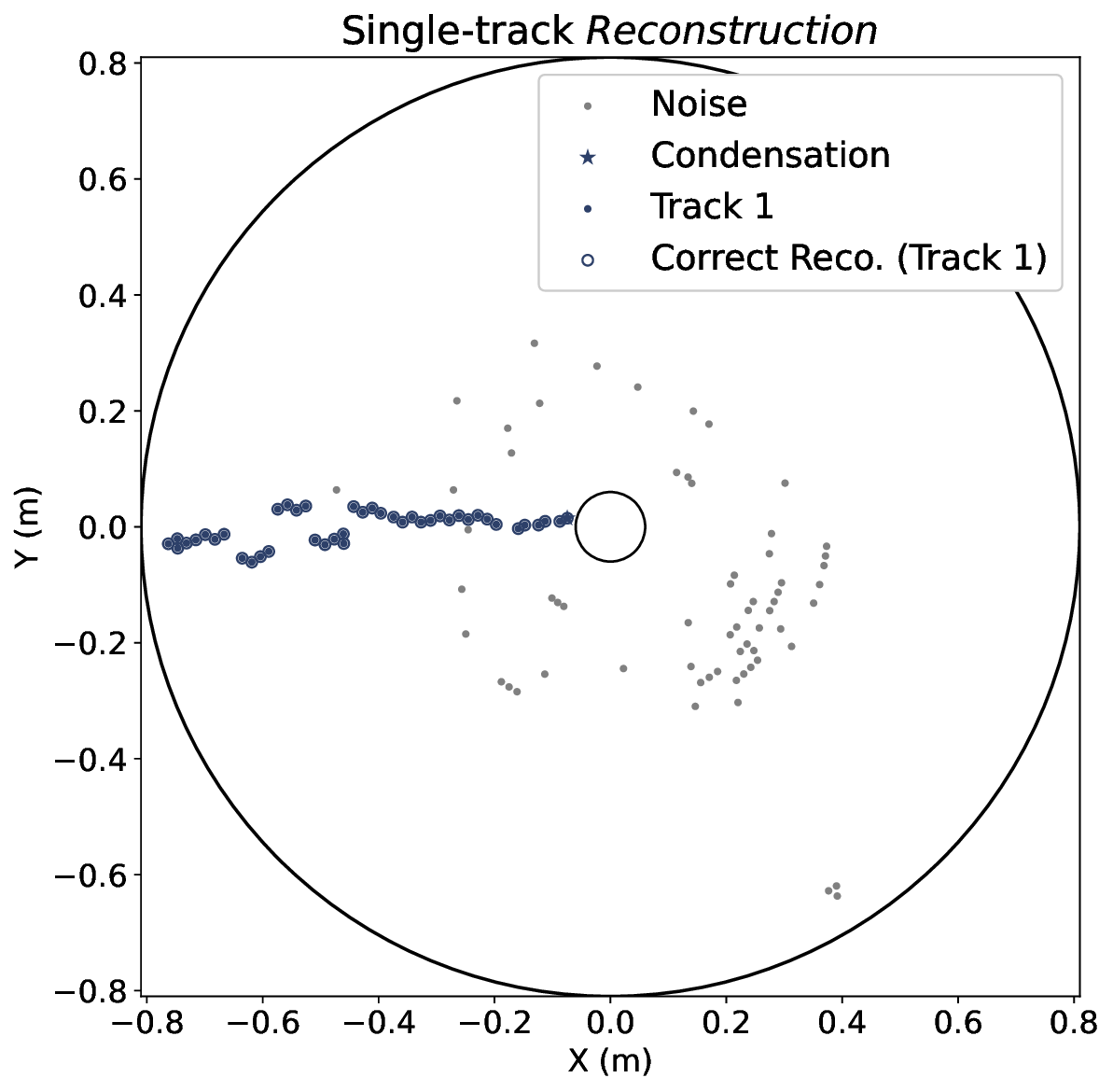}
    \end{minipage}
    \hfill
    \begin{minipage}[t]{0.31\textwidth}
        \centering
        \includegraphics[width=\linewidth]{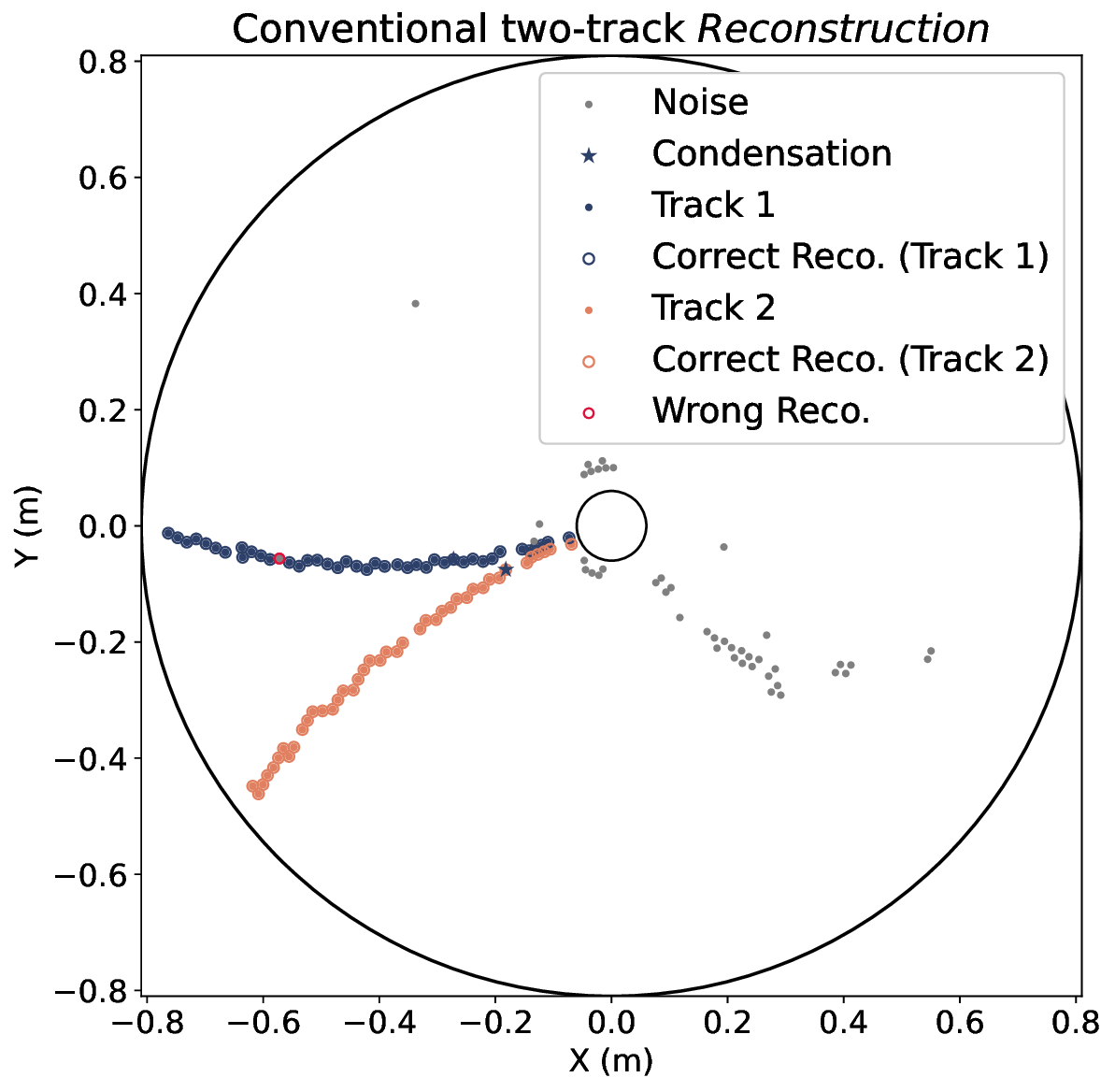}
    \end{minipage}
    \hfill
    \begin{minipage}[t]{0.31\textwidth}
        \centering
        \includegraphics[width=\linewidth]{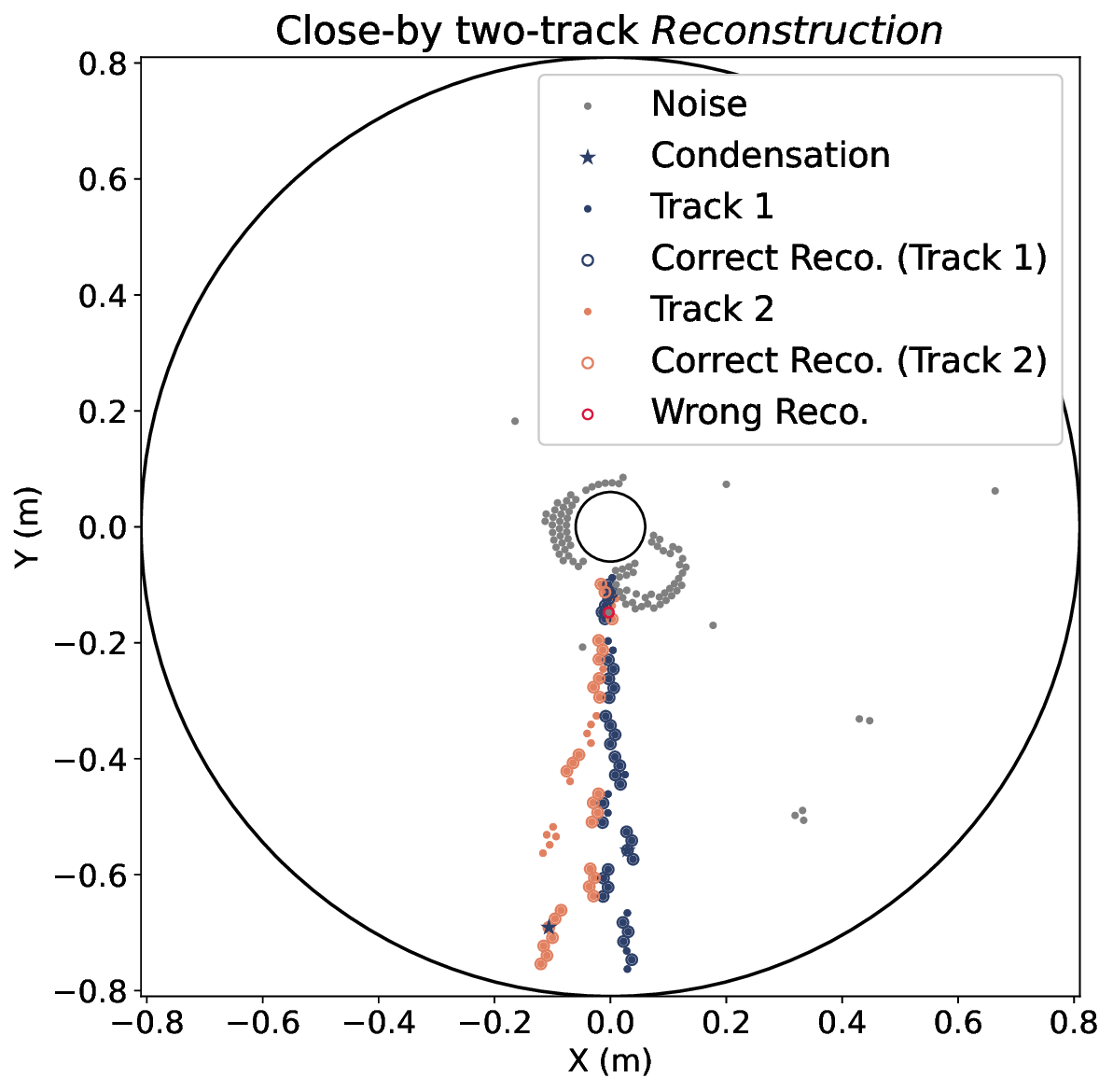}
    \end{minipage}
    
    \caption{Displays of reconstructed events in the x-y plane for a \textit{single-track} event (left), a \textit{conventional two-track} event (middle) and a \textit{close-by two-track} event (right). A condensation point on the track provides estimates the track parameters. This concept is closely related to the GNN finding method we use.}
    \label{fig:rec_event_display}
\end{figure}

Example displays of track reconstruction events are shown in figure~\ref{fig:rec_event_display} for different event categories.

%rec event display

\subsubsection{Hit efficiency and hit purity} 
\label{sec:hit_efficiency_purity}

The hit efficiency ($\epsilon_{\text{hit}}$) and hit purity ($p_{\text{hit}}$) for tracks found by both the \gnnfinder and the \baselinefinder are summarized in table~\ref{table:hit_performance_part}.

\begin{figure}[!htbp]
    \centering
    % 第一行：Hit efficiency 子图 (a)(b)
    % 子图a：Hit efficiency vs p_T^MC
    \begin{minipage}[t]{0.49\textwidth}
        \centering
        \includegraphics[width=\linewidth]{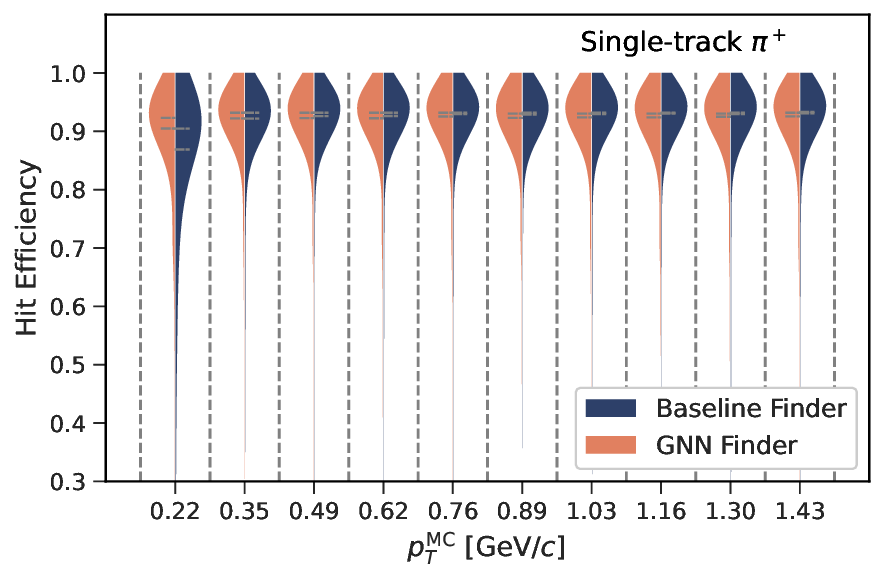}
    \end{minipage}
    \hfill % 列间均匀分隔
    % 子图b：Hit efficiency vs cosθ^MC
    \begin{minipage}[t]{0.49\textwidth}
        \centering
        \includegraphics[width=\linewidth]{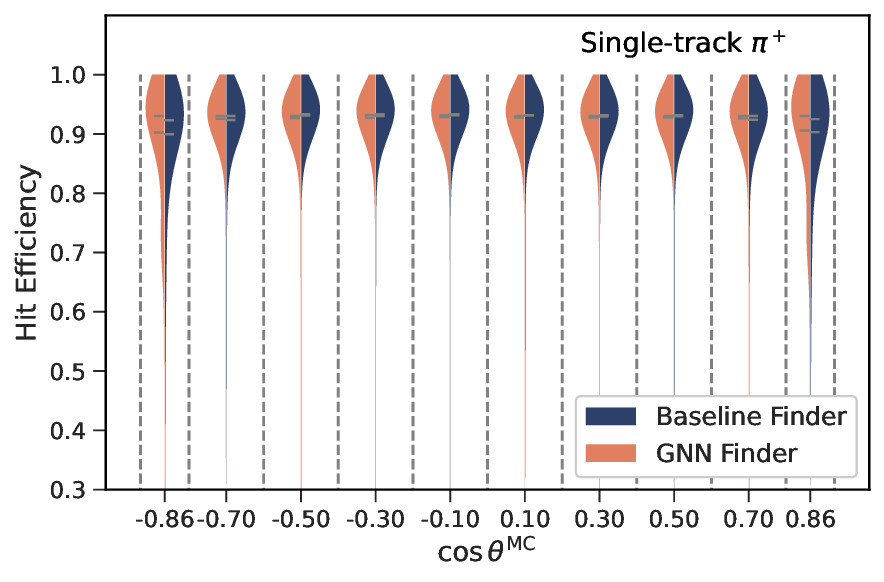}
    \end{minipage}
    
    % 第二行：Hit purity 子图 (c)(d)
    % 子图c：Hit purity vs p_T^MC
    \begin{minipage}[t]{0.49\textwidth}
        \centering
        \includegraphics[width=\linewidth]{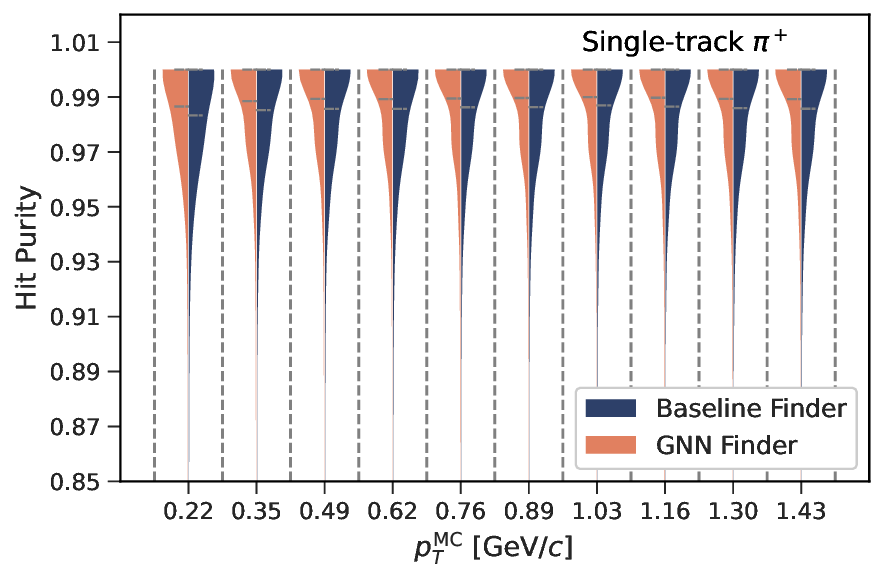}
    \end{minipage}
    \hfill % 列间均匀分隔
    % 子图d：Hit purity vs cosθ^MC
    \begin{minipage}[t]{0.49\textwidth}
        \centering
        \includegraphics[width=\linewidth]{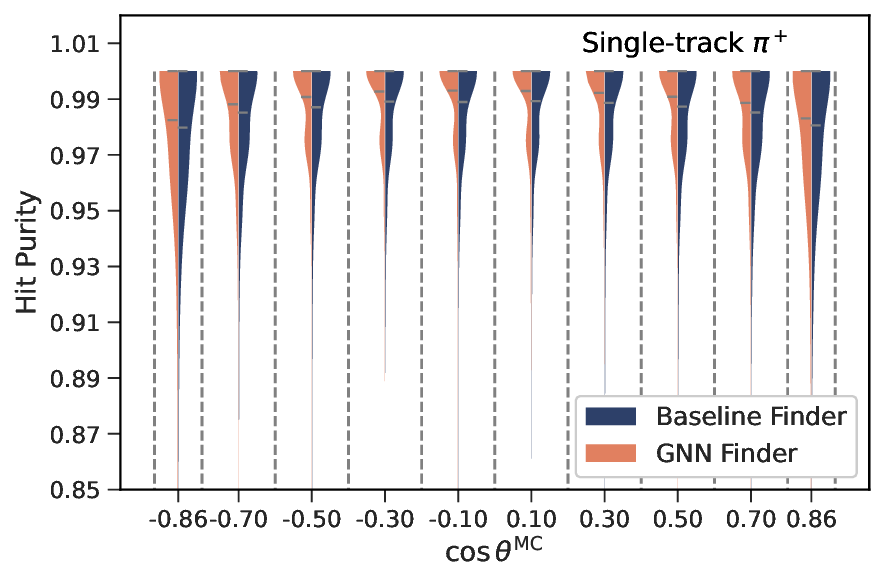}
    \end{minipage}

    % 整体标题：涵盖所有四张子图，明确对应关系
    \caption{Hit efficiency and hit purity for tracks found by both the \gnnfinder and the \textit{Baseline Finder}. Results are shown as functions of $p_{\mathrm{T}}^{\mathrm{MC}}$ (left column) and $\cos\theta^{\mathrm{MC}}$ (right column) for \textit{single-track} $\pi^+$ events.}
    \label{fig:hit_efficiency_purity_pi}
\end{figure}

For \textit{single-track} $\pi^+$ and \textit{conventional two-track} $\pi^+\pi^-$ events, the \gnnfinder exhibits comparable hit efficiency and hit purity to that of the \textit{Baseline Finder}. Figure~\ref{fig:hit_efficiency_purity_pi} shows the hit efficiency and hit purity as functions of $p_{\mathrm{T}}^{\mathrm{MC}}$ and $\cos\theta^{\mathrm{MC}}$ for \textit{single-track} events, comparing the performance of the \textit{GNN Finder} and the \textit{Baseline Finder}. The corresponding distributions for the \textit{conventional two-track} events are provided in the appendix~\ref{app:hit_efficiency_and_hit_purity}.

In contrast, for \textit{close-by two-track} events, the \gnnfinder exhibits a significant degradation in hit efficiency, while hit purity remains comparable. As this experiment represents an initial exploration, we anticipate that future investigations by researchers will further refine ML methods to address such scenarios. The corresponding distributions for \textit{close-by two-track} events are presented in appendix \ref{app:hit_efficiency_and_hit_purity}.

\begin{table}[!htbp]
    \centering
    \setlength{\belowcaptionskip}{6pt}  
    % 保持caption上方间距为默认值（和MDC表格一致）
    \setlength{\abovecaptionskip}{10pt}
    \caption{Hit efficiency ($\epsilon_{\text{hit}}$) and hit purity ($p_{\text{hit}}$) for different event categories.}
    \label{table:hit_performance_part}
    \small % 匹配jinstpub的表格字体大小
    \renewcommand{\arraystretch}{1.8} % 增大行间距，提升可读性
    \begin{tabular}{@{}l l l l l@{}} 
        \toprule[1.2pt] % 顶部粗黑线
        \textbf{in \%} & \textbf{Event Type} & $\boldsymbol{\epsilon_{\text{hit}}}$ & $\boldsymbol{p_{\text{hit}}}$  \\
        \midrule[0.8pt] % 表头与内容间细横线
        % 第一组：single-track $\pi^+$
        Baseline Finder 
        & Single-track ($\pi^+$) 
        & $92.24\substack{+0.12\\-0.12}$  
        & $98.58\substack{+0.05\\-0.05}$\\
        GNN Finder      
        & & $92.20\substack{+0.12\\-0.12}$  
        & $98.91\substack{+0.05\\-0.05}$\\
        % 第一组：single-track $\pi^-$
        % Baseline Finder 
       
        % \midrule[0.4pt] % 组间分隔细虚线（提升层次感）
        % 第三组：conventional two-track
        Baseline Finder 
        & Conventional two-track ($\pi^+\pi^-$) 
        & $90.87\substack{+0.14\\-0.14}$ 
        & $97.93\substack{+0.07\\-0.07}$  \\
        GNN Finder      
        & &$91.62\substack{+0.13\\-0.13}$ 
        & $98.83\substack{+0.05\\-0.05}$  \\
        % 第二组：close-by two-track
        Baseline Finder 
        & Close-by two-track ($\pi^+\pi^-$) 
        & $91.26\substack{+0.16\\-0.16}$ 
        & $97.95\substack{+0.08\\-0.08}$  \\
        GNN Finder      
        & &$82.68\substack{+0.21\\-0.21}$ 
        & $97.89\substack{+0.08\\-0.08}$  \\
        % \midrule[0.4pt] % 组间分隔细虚线
        \bottomrule[1.2pt] % 底部粗黑线
    \end{tabular}
    % 恢复默认行间距
    \renewcommand{\arraystretch}{1.0}
    
\end{table}

\subsubsection{Track finding and fitting efficiencies} 
\label{sec:track_efficiency}
The track finding and fitting efficiencies for tracks found by both the \gnnfinder and the \baselinefinder are summarized in table~\ref{table:performance_combined}.

For \textit{single-track} $\pi^+$ and \textit{conventional two-track} $\pi^+\pi^-$ events, we find that the \gnnfinder achieves track finding efficiencies comparable to that of the \textit{Baseline Finder}, with fitting efficiencies marginally lower, while its wrong charge rate is slightly higher. The track efficiency and track charge efficiency for \textit{single-track} $\pi^+$ events are shown as functions of $p_\mathrm{T}^{\mathrm{MC}}$ and $\cos\theta^{\mathrm{MC}}$ in figure~\ref{fig:track_eff_pi}; for \textit{conventional two-track} events, corresponding distribution figures are presented in appendix~\ref{app:track_finding_and_fitting_efficiency}.

\begin{table}[!htbp]
    \centering
    \setlength{\belowcaptionskip}{6pt}  
    \setlength{\abovecaptionskip}{10pt}
    \caption{Track finding and fitting efficiencies for different event categories.}
    \label{table:performance_combined}
    \small
    \renewcommand{\arraystretch}{1.8}
    \begin{tabular}{@{}l l l l l l@{}}
        \toprule[1.2pt]
         \textbf{in \%} & $\epsilon_{\text{track}}$     & $\epsilon_{\text{track,q}}$ & $R_{\text{clone}}$ & $R_{\text{fake}}$  & $R_{\text{wrong,q}}$ \\
        \midrule[0.8pt]
        \multicolumn{6}{@{}l}{\textbf{\textit{Single-track} $\pi^+$ events}} \\
        \midrule[0.4pt]
        Baseline Finder                  
        & $99.71\substack{+0.02\\-0.02}$ 
        & $99.69\substack{+0.02\\-0.02}$ 
        & $0.07\substack{+0.01\\-0.01}$ 
        & $0.01$                     
        & $0.02\substack{+0.01\\-0.01}$ \\
        GNN Finder                       
        & $99.81\substack{+0.02\\-0.02}$ 
        & $99.55\substack{+0.03\\-0.03}$ 
        & $0.00$                     
        & $0.01$                     
        & $0.27\substack{+0.02\\-0.02}$ \\
        Baseline Fitter                  
        & $99.70\substack{+0.02\\-0.02}$ 
        & $99.68\substack{+0.03\\-0.03}$ 
        & $0.06\substack{+0.01\\-0.01}$ 
        & $0.01$                     
        & $0.02\substack{+0.01\\-0.01}$ \\
        GNN Fitter                       
        & $99.75\substack{+0.02\\-0.02}$ 
        & $99.50\substack{+0.03\\-0.03}$ 
        & $0.00$                     
        & $0.01$                     
        & $0.25\substack{+0.02\\-0.02}$ \\
        \midrule[0.8pt]
        \multicolumn{6}{@{}l}{\textbf{\textit{Conventional two-track} $\pi^+\pi^-$ events}} \\
        \midrule[0.4pt]
        Baseline Finder                  & $99.63\substack{+0.03\\-0.03}$ & $99.59\substack{+0.03\\-0.03}$ & $0.10\substack{+0.01\\-0.01}$ & $0.01\substack{+0.01\\-0.01}$ & $0.04\substack{+0.01\\-0.01}$ \\
        GNN Finder                       & $99.50\substack{+0.03\\-0.03}$ & $99.31\substack{+0.04\\-0.04}$ & $0.00$                     & $0.02\substack{+0.01\\-0.01}$ & $0.19\substack{+0.02\\-0.02}$ \\
        Baseline Fitter                  & $99.62\substack{+0.03\\-0.03}$ & $99.59\substack{+0.03\\-0.03}$ & $0.10\substack{+0.01\\-0.01}$ & $0.01\substack{+0.01\\-0.01}$ & $0.03\substack{+0.01\\-0.01}$ \\
        GNN Fitter                       & $99.45\substack{+0.04\\-0.04}$ & $99.29\substack{+0.04\\-0.04}$ & $0.00$                     & $0.02\substack{+0.01\\-0.01}$ & $0.16\substack{+0.02\\-0.02}$ \\
        \midrule[0.8pt]
        \multicolumn{6}{@{}l}{\textbf{\textit{Close-by two-track} $\pi^+\pi^-$ events}} \\
        \midrule[0.4pt]
        Baseline Finder                  & $99.55\substack{+0.03\\-0.03}$ & $99.52\substack{+0.03\\-0.03}$ & $0.13\substack{+0.02\\-0.02}$ & $0.02\substack{+0.01\\-0.01}$ & $0.03\substack{+0.01\\-0.01}$ \\
        GNN Finder                       & $76.22\substack{+0.20\\-0.20}$ & $75.44\substack{+0.21\\-0.21}$ & $0.14\substack{+0.02\\-0.02}$ & $0.32\substack{+0.03\\-0.03}$ & $0.77\substack{+0.04\\-0.04}$ \\
        Baseline Fitter                  & $99.53\substack{+0.02\\-0.02}$ & $99.50\substack{+0.03\\-0.03}$ & $0.12\substack{+0.02\\-0.02}$ & $0.01$                     & $0.03\substack{+0.01\\-0.01}$ \\
        GNN Fitter                       & $75.85\substack{+0.20\\-0.20}$ & $75.27\substack{+0.21\\-0.21}$ & $0.12\substack{+0.02\\-0.02}$ & $0.20\substack{+0.02\\-0.02}$ & $0.58\substack{+0.04\\-0.04}$ \\
        \bottomrule[1.2pt]
    \end{tabular}
    \renewcommand{\arraystretch}{1.0}
\end{table}

%Track efficiency and track charge efficiency
\begin{figure}[!htbp]
    \centering
    \renewcommand{\arraystretch}{1.2} % 调整子图上下间距，避免拥挤
    % 第1行（两列）
    \begin{minipage}[t]{0.48\textwidth} % 两列均分宽度，预留间距
        \centering
        \includegraphics[width=\linewidth]{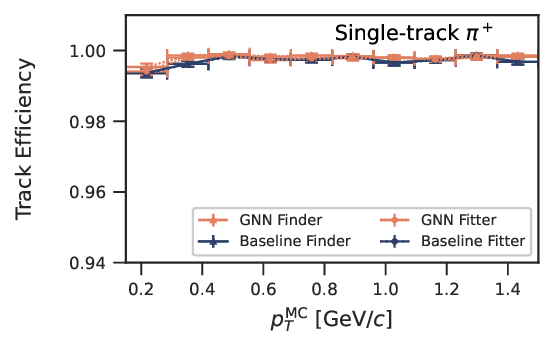} % 子图1(a)
    \end{minipage}
    \hfill % 两列之间留白，保持整洁
    \begin{minipage}[t]{0.48\textwidth}
        \centering
        \includegraphics[width=\linewidth]{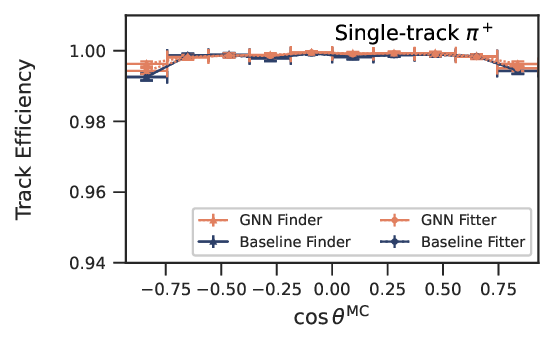} % 子图1(b)
    \end{minipage}
    
    % 第2行（两列），此处需替换为你另外两个子图的路径（替换占位符）
    \begin{minipage}[t]{0.48\textwidth}
        \centering
        \includegraphics[width=\linewidth]{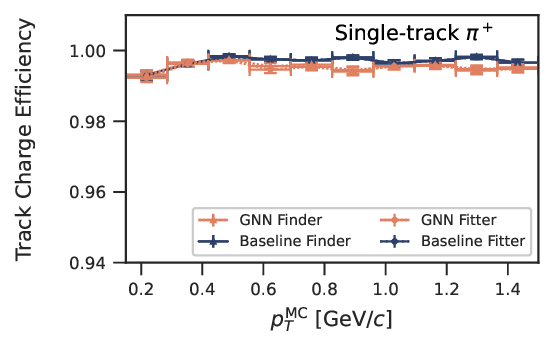}
    \end{minipage}
    \hfill
    \begin{minipage}[t]{0.48\textwidth}
        \centering
        \includegraphics[width=\linewidth]{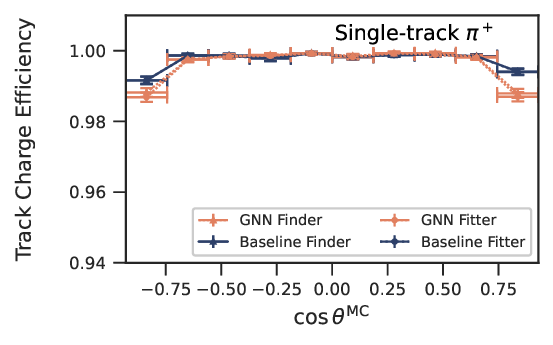}
    \end{minipage}
    
    \caption{Track efficiency and track charge efficiency for tracks found by both the \gnnfinder (orange) and the \baselinefinder (blue) with and without fitting. Results are shown as functions of $p_\mathrm{T}^{\mathrm{MC}}$ (left column) and $\cos\theta^{\mathrm{MC}}$ (right column) for \textit{single-track} $\pi^+$ events.}
    \label{fig:track_eff_pi}
\end{figure}

\newpage
In contrast, for \textit{close-by two-track} events, where the two tracks are closely spaced, the performance gap becomes pronounced. The \gnnfinder and \textit{Fitter} suffer a substantial drop in track finding and fitting efficiencies compared to the Baseline's near $\sim$100\% values. 
Additionally, the wrong charge rate rises sharply to 0.77\% (the \textit{GNN Finder}) and 0.58\% (the \textit{GNN Fitter}), far exceeding the Baseline's $\sim$0.03\%. The corresponding distribution figures are presented in appendix \ref{app:track_finding_and_fitting_efficiency}.

\subsubsection{Track parameter performance}
\label{sec:resolution}

\begin{figure}[htbp]
    \centering  % 图片居中
    \includegraphics[width=0.6\textwidth]{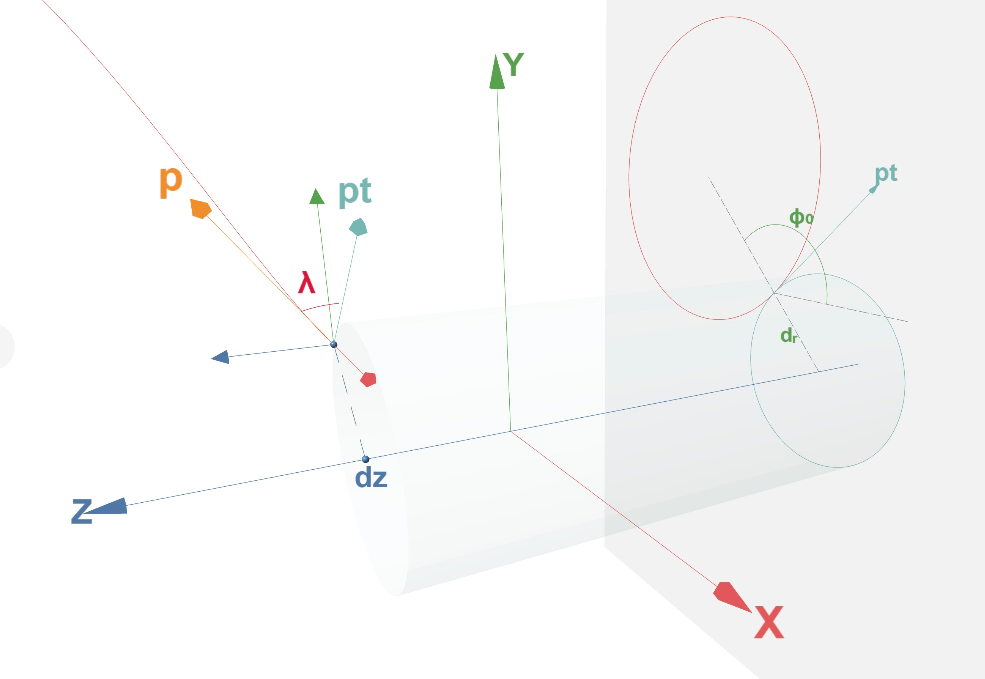} 
    \caption{Helix parametrization of a particle trajectory.}
    \label{fig:helix} % 设置标签，用于文中引用
\end{figure}

The trajectory for a charged track in a uniform magnetic field can be represented by a helix which can be defined with five track parameters \((d_r, \phi_0, \kappa, d_z, \tan\lambda)^T\), defined at the POCA to the origin. An animated visualization of this helix parametrization for a particle trajectory is available at \url{https://lyqian1220.github.io/}; a static illustration is shown in figure~\ref{fig:helix}.The five track parameters are defined as follows:

\begin{itemize}
    \item $d_r$ is the signed distance from the POCA to the origin in the $x$-$y$ plane (in cm). The sign is defined by $(\vec{d} \times \vec{p})$, where $\vec{d}$ is the vector from origin to the track and $\vec{p}$ is the tangent to the track direction.
    
    \item $\phi_0$  is the azimuthal angle of the POCA relative to the helix center in the transverse plane (in rad). The range of $\phi_0$ is from $0$ to $2\pi$.
    
    \item $\kappa$ is the reciprocal of the transverse momentum $p_T$ (in $(\text{GeV}/c)^{-1}$). The sign of $\kappa$ represents the charge of the track.
    
    \item $d_z$ is the z-coordinate of the POCA relative to the origin (in cm).
    
    \item $\tan\lambda$ is the slope of the track, or the tangent of the dip angle $\lambda$. The polar angle of the track is defined as $\theta \equiv \pi/2 - \lambda$.
\end{itemize}

\begin{figure}[h!tbp]
  \centering
  % 三张图并列，每张宽度设为 0.3\textwidth（总和 < 1 避免溢出）
  \includegraphics[width=0.3\textwidth]{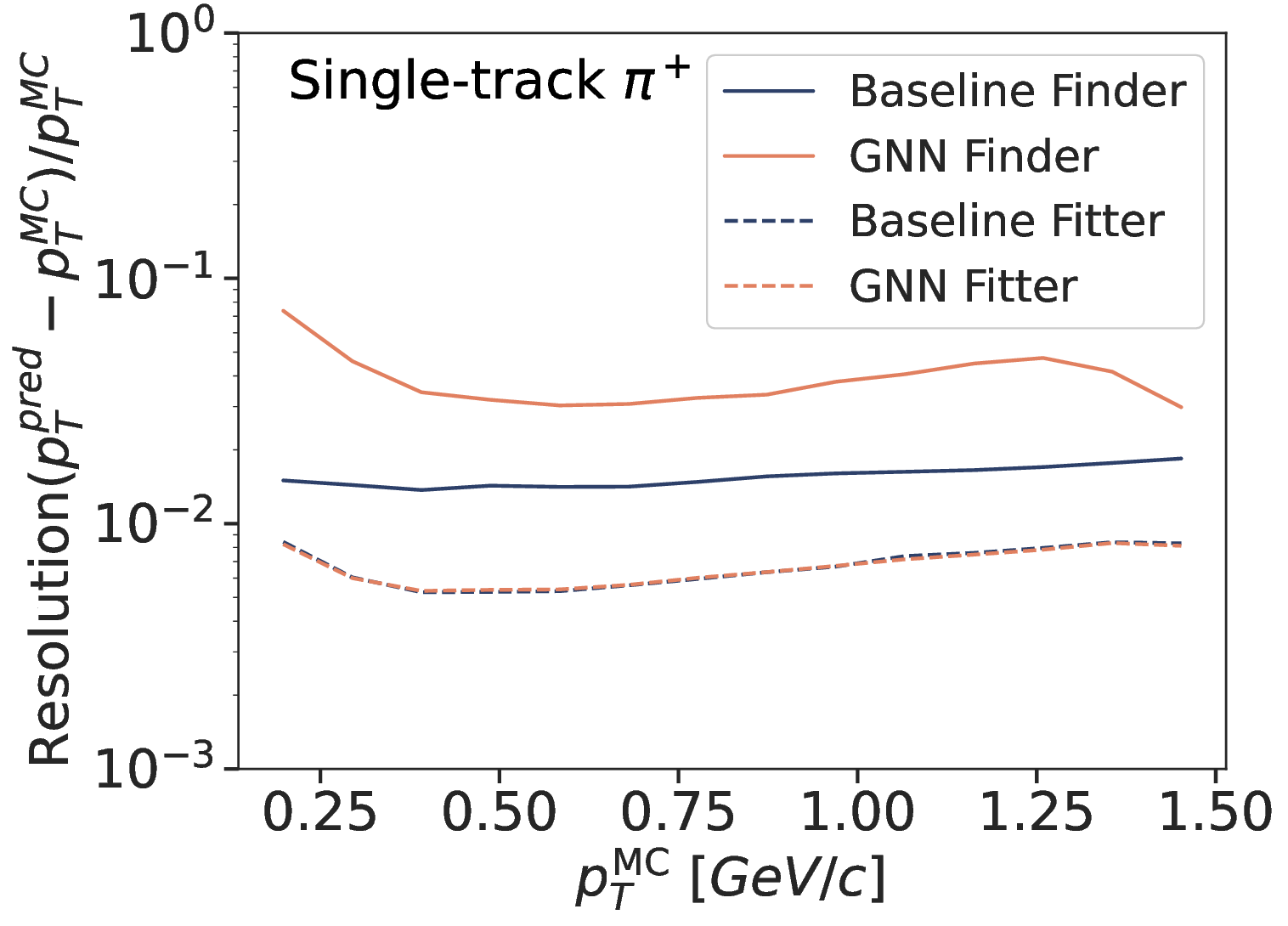}
  \hfill  % 图间均匀分布
  \includegraphics[width=0.3\textwidth]{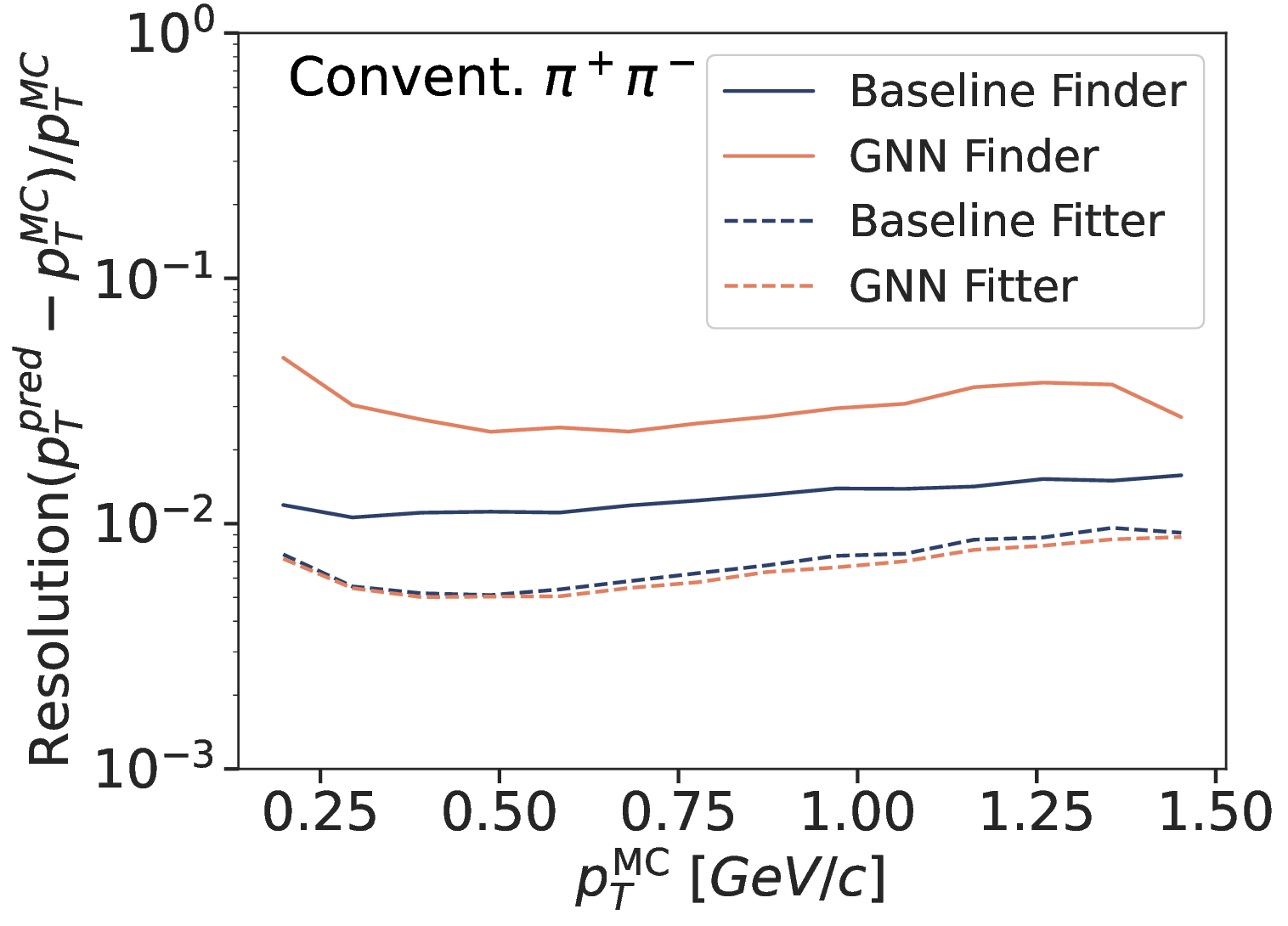}
  \hfill
  \includegraphics[width=0.3\textwidth]{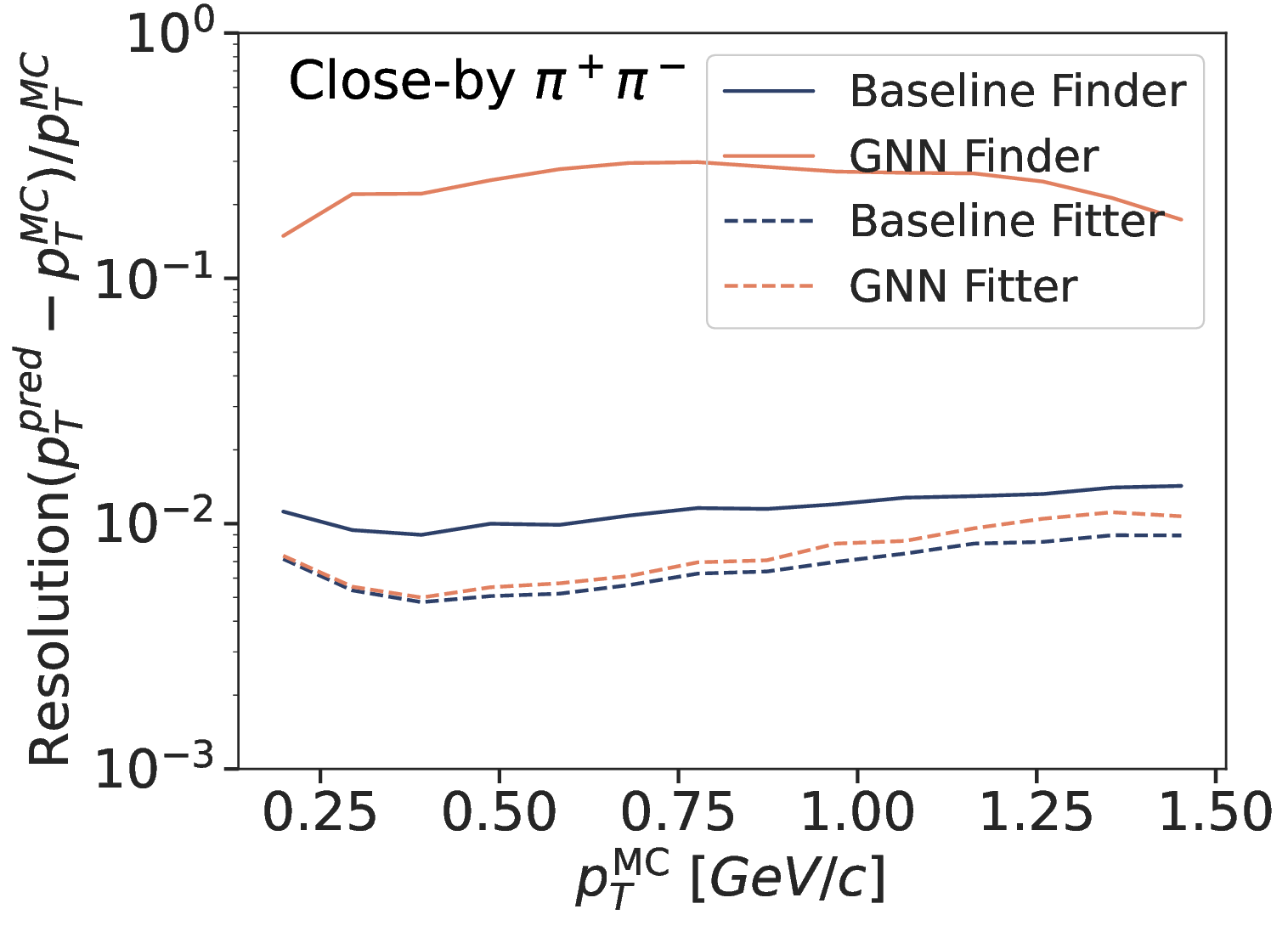}
  \caption{Relative transverse momentum resolution for tracks found by both the \gnnfinder and the \textit{Baseline Finder} with and without fitting. Results are shown as a function of $p_\mathrm{T}^{\mathrm{MC}}$ for \textit{single-track} events (left), \textit{conventional two-track} events (middle) and \textit{close-by two-track} (right) events.}
  \label{fig:pt_resolution}
\end{figure}

We evaluate the transverse momentum resolution for reconstructed tracks with correct charge that are found by both the \gnnfinder and the \textit{Baseline Finder}. The relative transverse momentum resolution of the \gnnfinder is inferior to that of the \textit{Baseline Finder}, while the \gnnfitter is comparable to the \textit{Baseline Fitter}, as shown in figure~\ref{fig:pt_resolution}.
However, in the case of \textit{close-by two-track} events, the \gnnfinder exhibits clearly worse resolution. The detailed distributions of track parameters for both the \gnnfinder and the \baselinefinder with and without fitting are shown in appendix \ref{app:track_par}.

\newpage
\section{Conclusion}
This work presents an open dataset for ML-based track reconstruction, built from realistic drift chamber simulation and detector response. It covers the phase space within the detector acceptance, including \textit{single-track}, \textit{conventional two-track} and \textit{close-by two-track} events with realistic noise overlay. Additionally, we establish a set of evaluation metrics and complete benchmark experiments using this dataset. Preliminary results show that while the \textit{GNN Finder} achieves track reconstruction performance comparable to that of the \textit{Baseline Finder} for \textit{single-track} and \textit{conventional two-track} events, its performance degrades significantly when handling \textit{close-by two-track} events.

In conclusion, this work addresses the shortage of drift chamber track reconstruction datasets and provides specific evaluation metrics for fair and reproducible comparison for the ML-based tracking methods, thereby hoping to promote advancements and innovations in the field.

\section{Outlook}

Future work will first focus on improving the dataset in richness and applicability. It will be extended to include both MC simulation and real data, covering displaced tracks, curved tracks and physics events. In addition, a sample covering both the inner tracker and the drift chamber will be provided. This will significantly enrich the data diversity and make the dataset more representative of real experimental scenarios, ultimately improving reconstruction performance, rare-signal sensitivity and discovery potential. Moreover, evaluation of the baseline finding and fitting methods will be enabled via public interfaces; at present, access is available only upon request and formal coordination with our team.

\newpage
\appendix

\section{Hit efficiency and hit purity}
\label{app:hit_efficiency_and_hit_purity}

\begin{figure}[!htbp]
    \centering
    % 第一行：Hit efficiency 子图 (a)(b)
    \begin{minipage}[t]{0.49\textwidth}
        \centering
        \includegraphics[width=\linewidth]{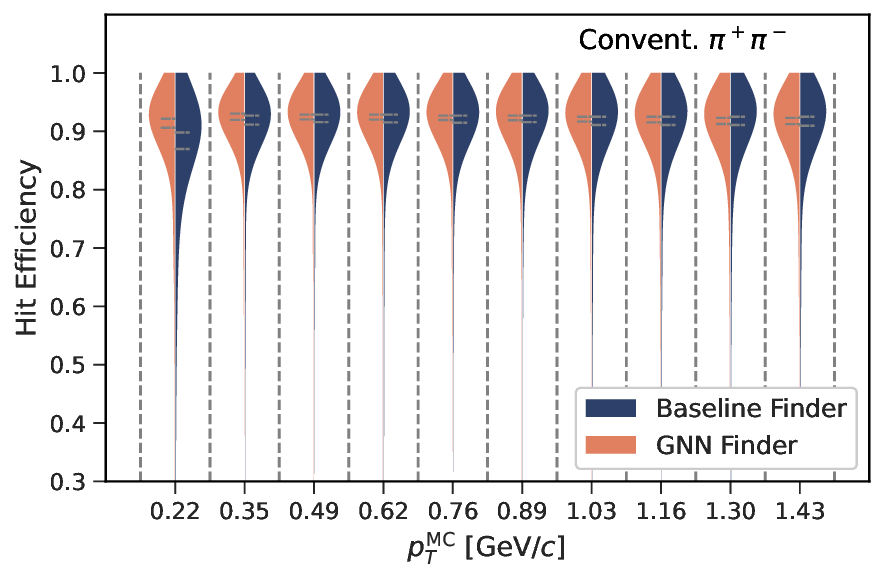}
    \end{minipage}
    \hfill
    \begin{minipage}[t]{0.49\textwidth}
        \centering
        \includegraphics[width=\linewidth]{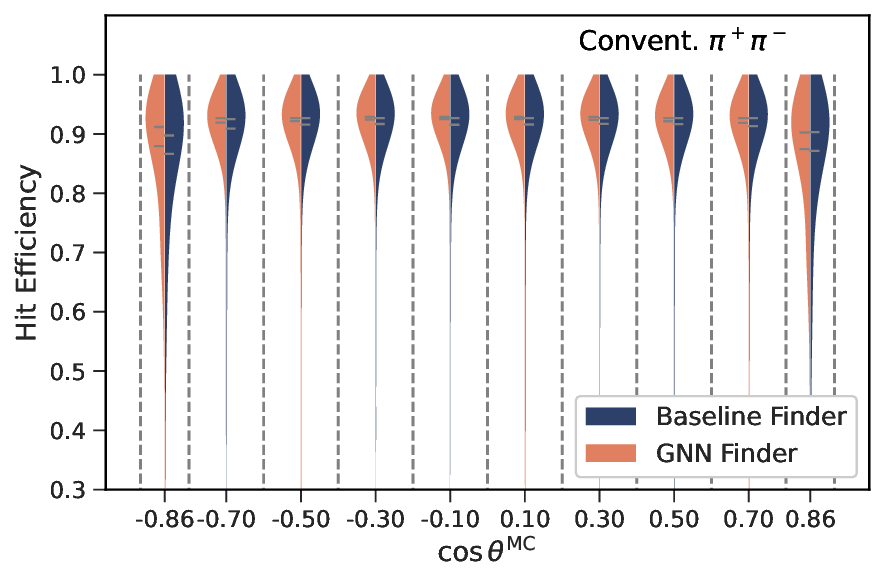}
    \end{minipage}
    
    % 第二行：Hit purity 子图 (c)(d)
    \begin{minipage}[t]{0.49\textwidth}
        \centering
        \includegraphics[width=\linewidth]{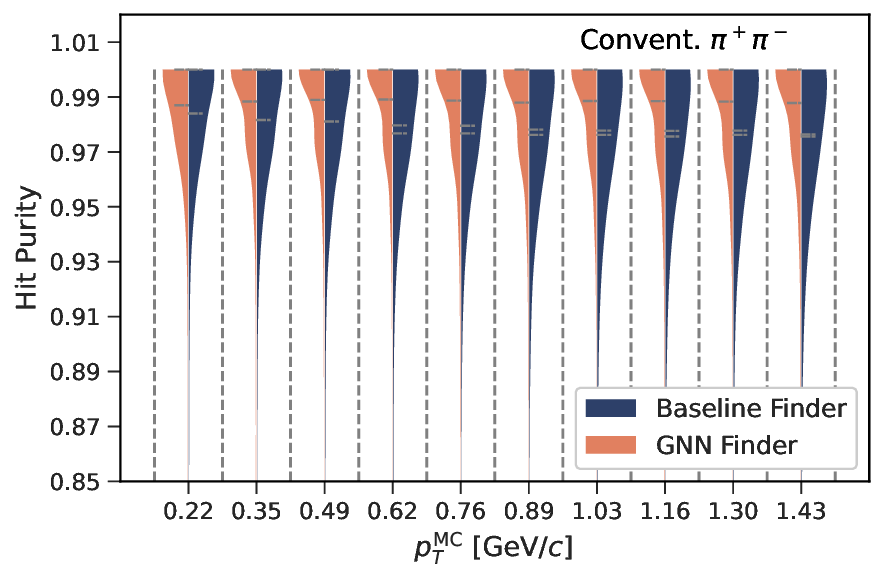}
    \end{minipage}
    \hfill
    \begin{minipage}[t]{0.49\textwidth}
        \centering
        \includegraphics[width=\linewidth]{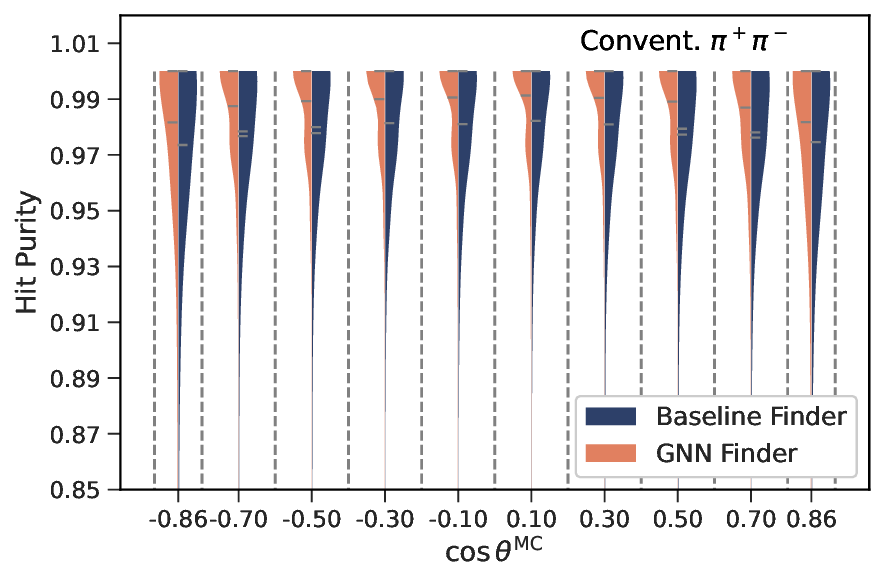}
    \end{minipage}

    \caption{Hit efficiency and hit purity for tracks found by both the \gnnfinder and the \textit{Baseline Finder}. Results are shown as functions of $p_{\mathrm{T}}^{\mathrm{MC}}$ (left column) and $\cos\theta^{\mathrm{MC}}$ (right column) for \textit{conventional two-track} $\pi^+\pi^-$ events.}
    \label{fig:hit_eff_pur_convent}
\end{figure}

\begin{figure}[!htbp]
    \centering
    % 第一行：Hit efficiency 子图 (a)(b)
    \begin{minipage}[t]{0.49\textwidth}
        \centering
        \includegraphics[width=\linewidth]{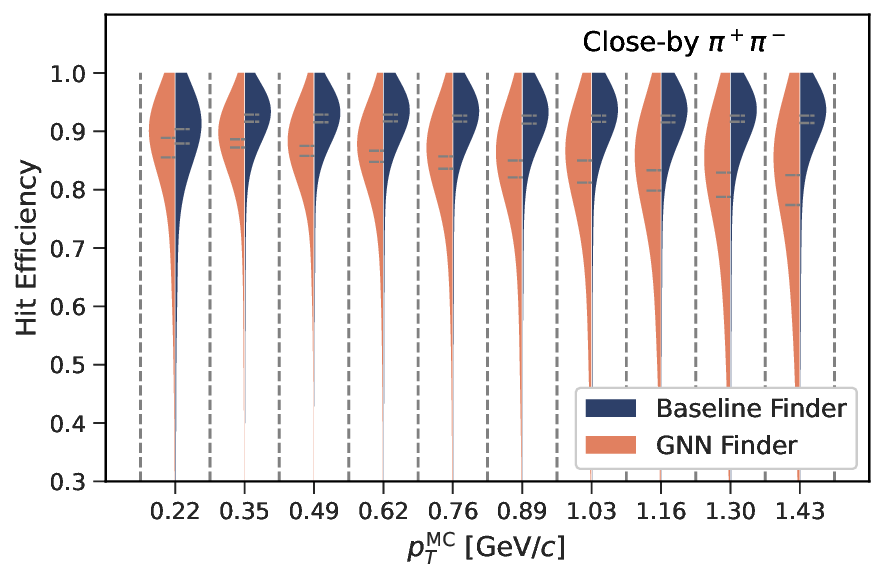}
    \end{minipage}
    \hfill % 列间均匀分隔
    \begin{minipage}[t]{0.49\textwidth}
        \centering
        \includegraphics[width=\linewidth]{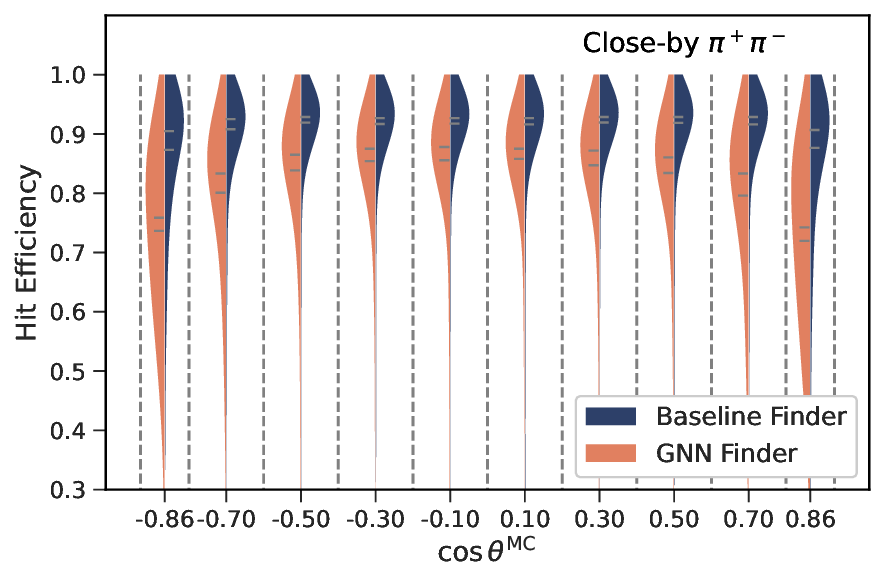}
    \end{minipage}

    % 第二行：Hit purity 子图 (c)(d)
    \begin{minipage}[t]{0.49\textwidth}
        \centering
        \includegraphics[width=\linewidth]{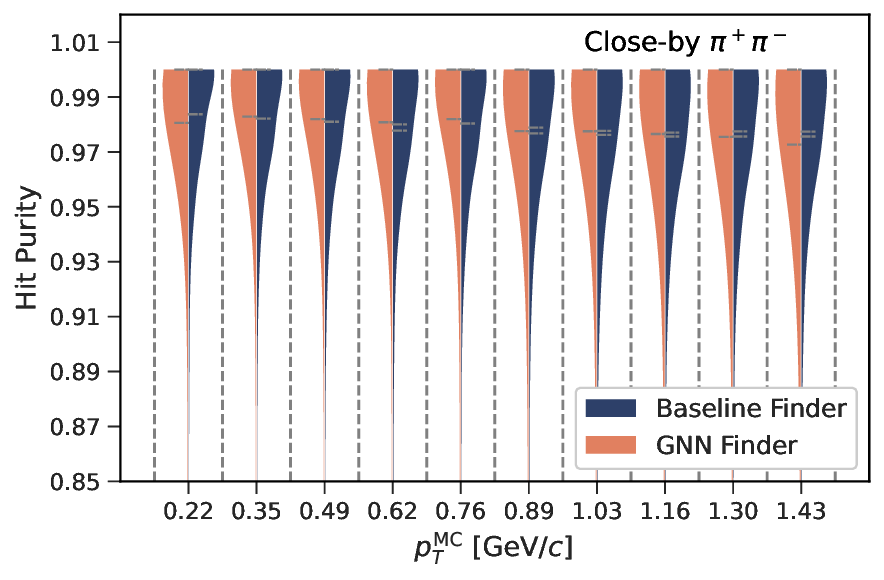}
    \end{minipage}
    \hfill % 列间均匀分隔
    \begin{minipage}[t]{0.49\textwidth}
        \centering
        \includegraphics[width=\linewidth]{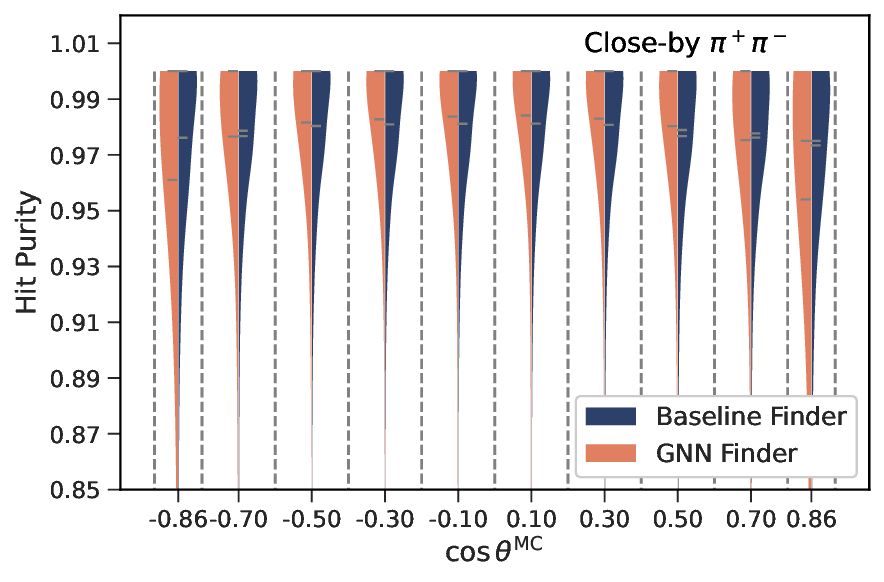}
    \end{minipage}

    % 整体标题：覆盖所有四张子图，明确实验条件
    \caption{Hit efficiency and hit purity for tracks found by both the \gnnfinder and the \textit{Baseline Finder}. Results are shown as functions of $p_{\mathrm{T}}^{\mathrm{MC}}$ (left column) and $\cos\theta^{\mathrm{MC}}$ (right column) for \textit{close-by two-track} $\pi^+\pi^-$ events.}
    \label{fig:hit_eff_pur_closeby}
\end{figure}

Figure~\ref{fig:hit_eff_pur_convent} shows the hit efficiency and hit purity for \textit{conventional two-track} $\pi^+\pi^-$ events.
For \textit{close-by two-track} $\pi^+\pi^-$ events, the performance is illustrated in figure~\ref{fig:hit_eff_pur_closeby}. Specifically, the hit efficiency of the \gnnfinder is mainly degraded by events with high transverse momentum and large polar angles.

% Conventional two-track 事件：Hit efficiency + Hit purity 2×2 子图

% \newpage

% Close-by two-track 事件：Hit efficiency + Hit purity 2×2 子图

\clearpage
\section{Track finding and fitting efficiencies}
\label{app:track_finding_and_fitting_efficiency}
\begin{figure}[!htbp]
    \centering
    \renewcommand{\arraystretch}{1.2} % 调整子图上下间距，避免拥挤
    % 第1行（两列）
    \begin{minipage}[t]{0.48\textwidth} % 两列均分宽度，预留间距
        \centering
        \includegraphics[width=\linewidth]{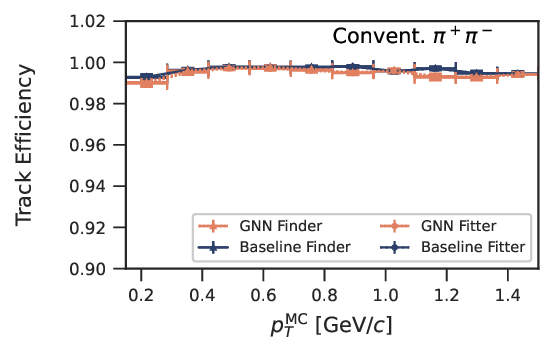} % 子图1(a)
    \end{minipage}
    \hfill % 两列之间留白，保持整洁
    \begin{minipage}[t]{0.48\textwidth}
        \centering
        \includegraphics[width=\linewidth]{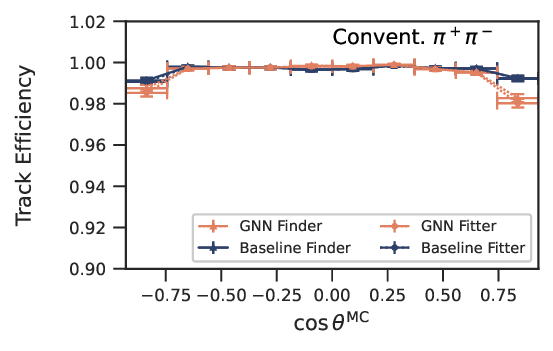} % 子图1(b)
    \end{minipage}
    
    % 第2行（两列），此处需替换为你另外两个子图的路径（替换占位符）
    \begin{minipage}[t]{0.48\textwidth}
        \centering
        \includegraphics[width=\linewidth]{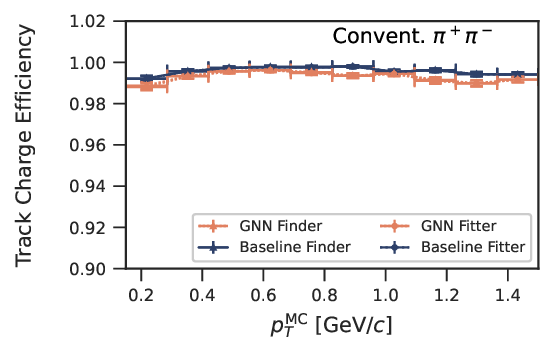}
    \end{minipage}
    \hfill
    \begin{minipage}[t]{0.48\textwidth}
        \centering
        \includegraphics[width=\linewidth]{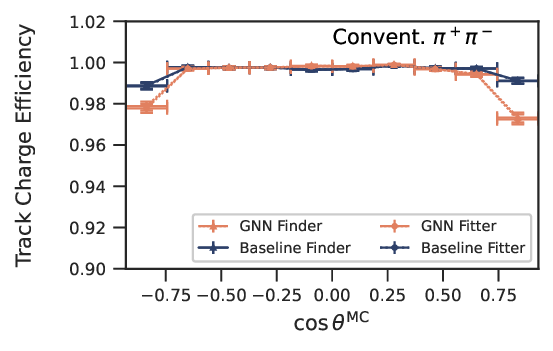}
    \end{minipage}
    
    \caption{Track efficiency and track charge efficiency for tracks found by both the \gnnfinder (orange) and the \baselinefinder (blue) with and without fitting. Results are shown as functions of $p_\mathrm{T}^{\mathrm{MC}}$ (left column) and $\cos\theta^{\mathrm{MC}}$ (right column) for \textit{conventional two-track} $\pi^+\pi^-$ events.}
    \label{fig:track_eff_convent}
\end{figure}

\begin{figure}[!htbp]
    \centering
    \renewcommand{\arraystretch}{1.2} % 调整子图上下间距，避免拥挤
    % 第1行（两列）
    \begin{minipage}[t]{0.48\textwidth} % 两列均分宽度，预留间距
        \centering
        \includegraphics[width=\linewidth]{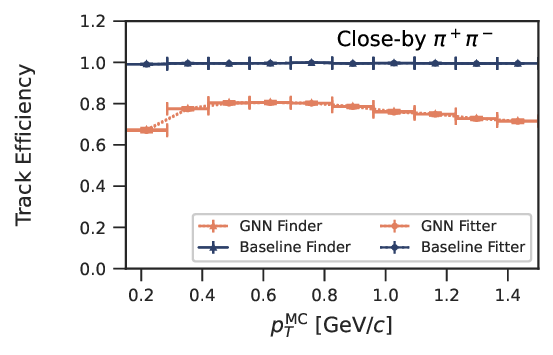} % 子图1(a)
    \end{minipage}
    \hfill % 两列之间留白，保持整洁
    \begin{minipage}[t]{0.48\textwidth}
        \centering
        \includegraphics[width=\linewidth]{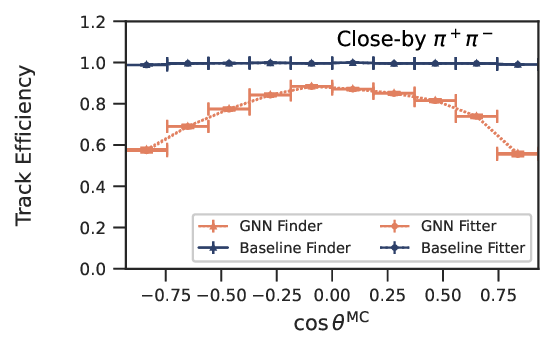} % 子图1(b)
    \end{minipage}
    
    % 第2行（两列），此处需替换为你另外两个子图的路径（替换占位符）
    \begin{minipage}[t]{0.48\textwidth}
        \centering
        \includegraphics[width=\linewidth]{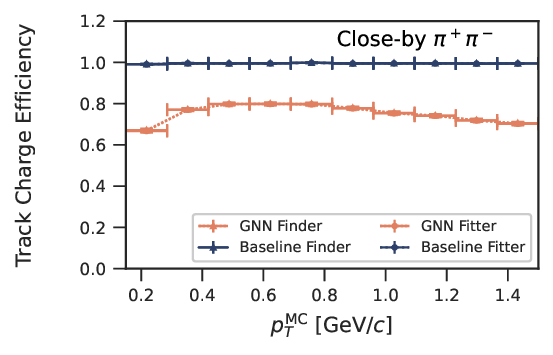}
    \end{minipage}
    \hfill
    \begin{minipage}[t]{0.48\textwidth}
        \centering
        \includegraphics[width=\linewidth]{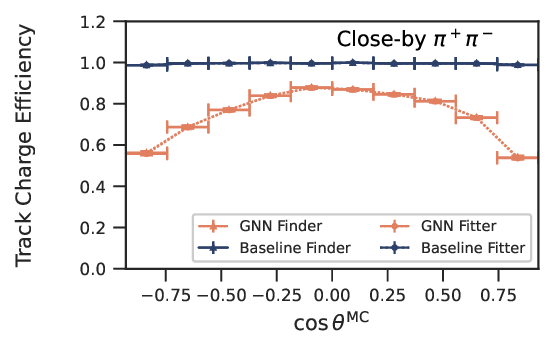}
    \end{minipage}
    
    \caption{Track efficiency and track charge efficiency for tracks found by both the \gnnfinder (orange) and the \baselinefinder (blue) with and without fitting. Results are shown as functions of $p_\mathrm{T}^{\mathrm{MC}}$ (left column) and $\cos\theta^{\mathrm{MC}}$ (right column) for \textit{close-by two-track} $\pi^+\pi^-$ events.}
    \label{fig:track_eff_closeby}
\end{figure}

Figure~\ref{fig:track_eff_convent} and \ref{fig:track_eff_closeby} show the track efficiency and track charge efficiency for the \gnnfinder and the \baselinefinder with and without fitting in \textit{conventional two-track} $\pi^+\pi^-$ and \textit{close-by two-track} events.

%Track efficiency and track charge efficiency

%Track efficiency and track charge efficiency

\clearpage
% \newpage
\section{Track parameters}
\label{app:track_par}

\begin{figure}[!htbp]
  \centering
  % 第一行带标注
  \begin{minipage}[t]{0.312\textwidth}
    \centering
    \includegraphics[width=\linewidth]{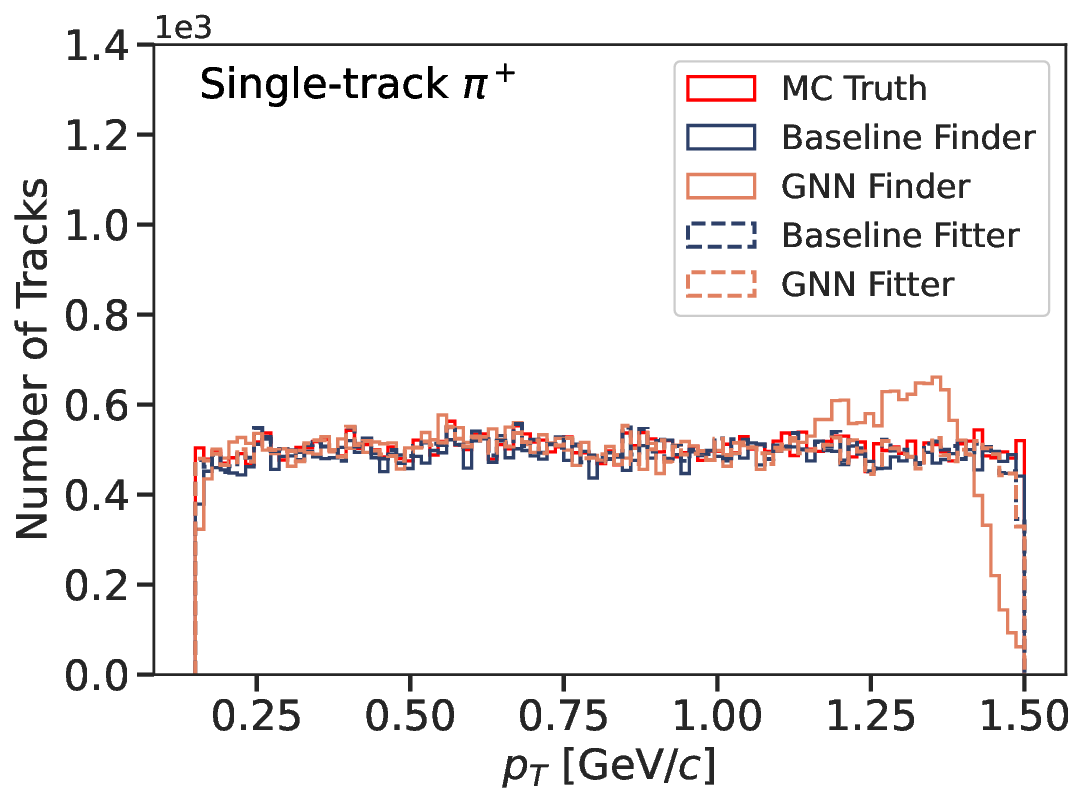}
  \end{minipage}
  \hfill
  \begin{minipage}[t]{0.3\textwidth}
    \centering
    \includegraphics[width=\linewidth]{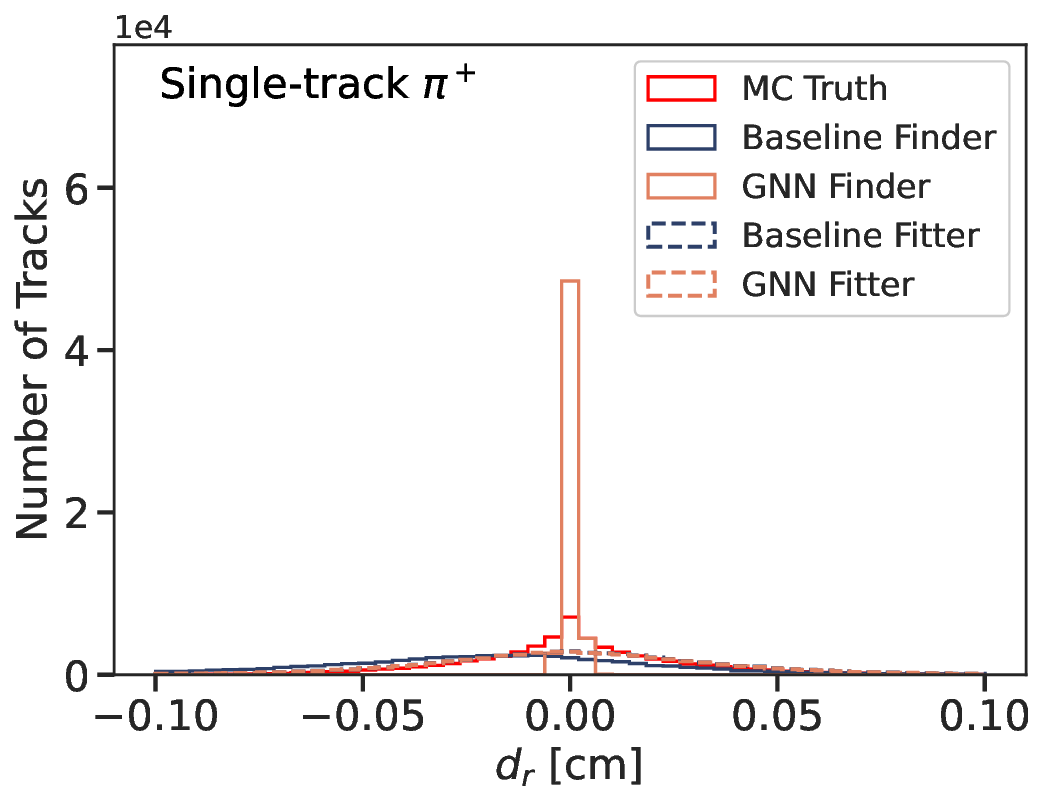}
  \end{minipage}
  \hfill
  \begin{minipage}[t]{0.31\textwidth}
    \centering
    \includegraphics[width=\linewidth]{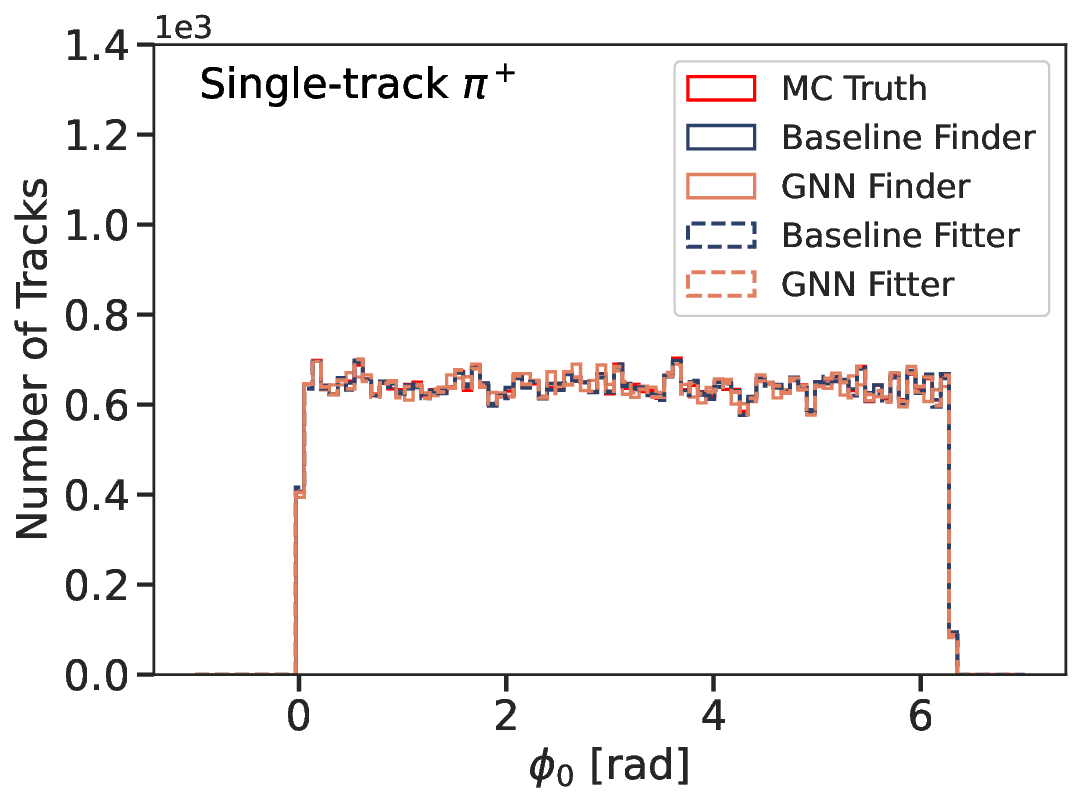}
  \end{minipage}
  
  \vspace{8pt}  % 带标注时可适当加大垂直间距
  
  % 第二行带标注
  \begin{minipage}[t]{0.31\textwidth}
    \centering
    \includegraphics[width=\linewidth]{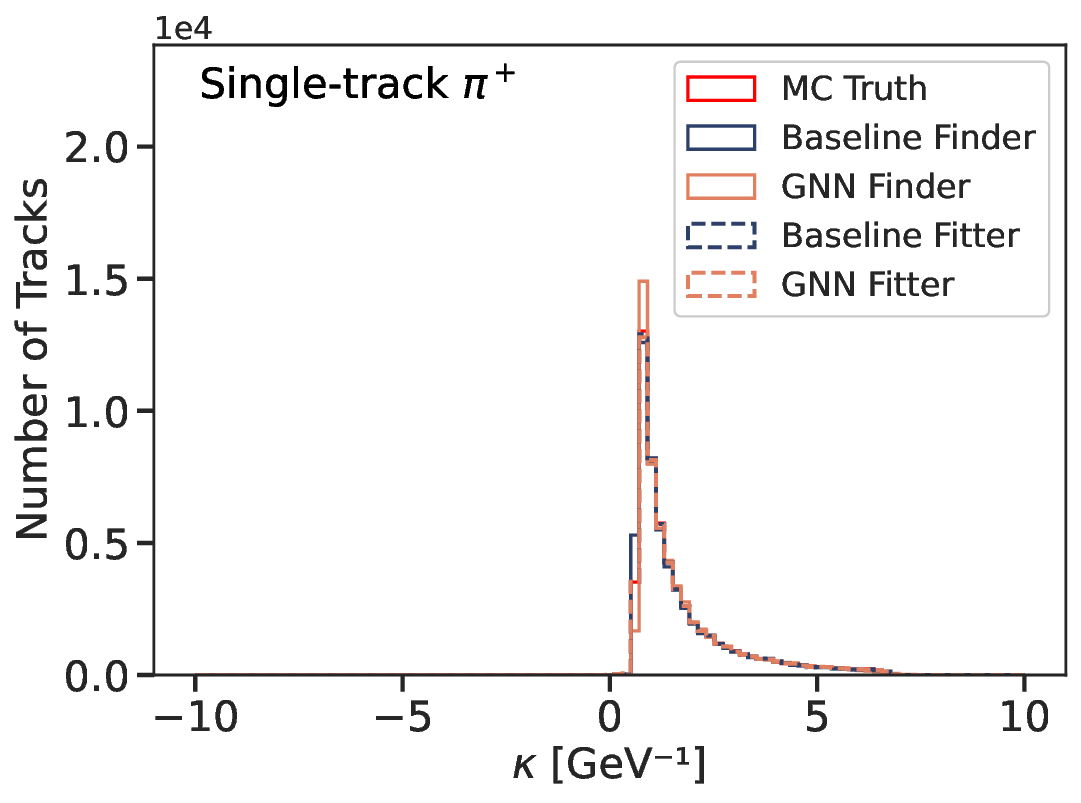}
  \end{minipage}
  \hfill
  \begin{minipage}[t]{0.3\textwidth}
    \centering
    \includegraphics[width=\linewidth]{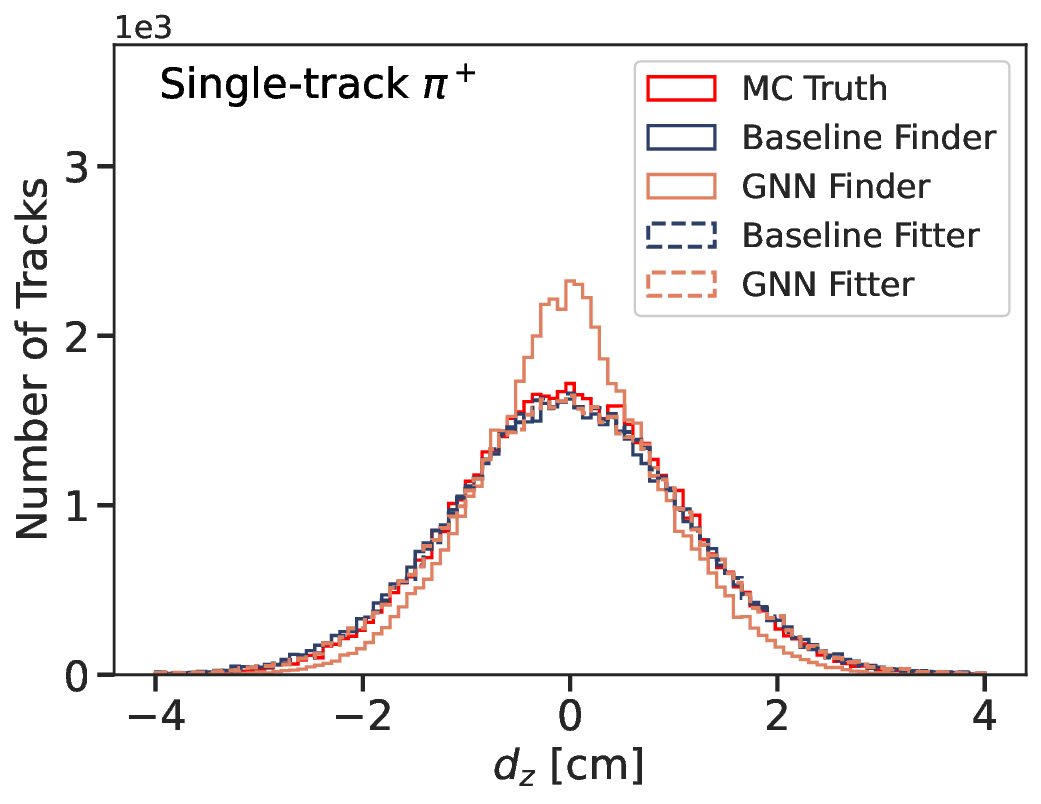}
  \end{minipage}
  \hfill
  \begin{minipage}[t]{0.31\textwidth}
    \centering
    \includegraphics[width=\linewidth]{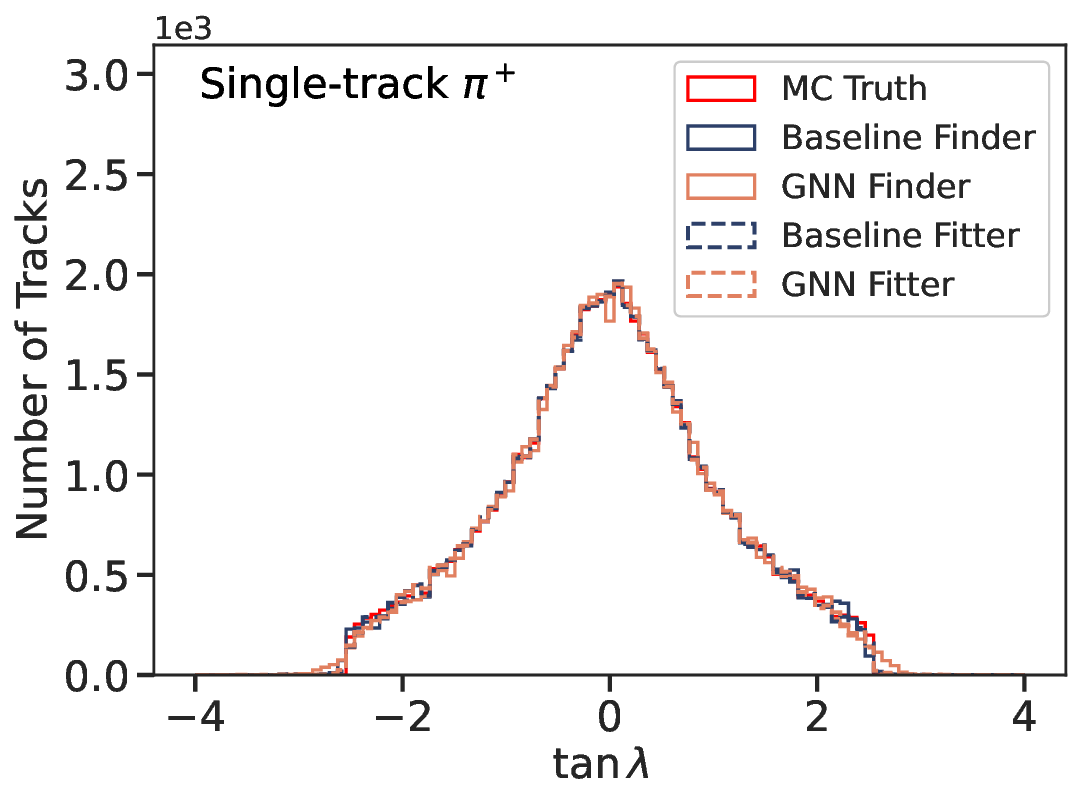}
  \end{minipage}

  \caption{Track parameters of \textit{single-track} $\pi^+$ events.}
  \label{fig:single_track_parameters}
\end{figure}

\begin{figure}[!htbp]
  \centering
  % 第一行带标注
  \begin{minipage}[t]{0.31\textwidth}
    \centering
    \includegraphics[width=\linewidth]{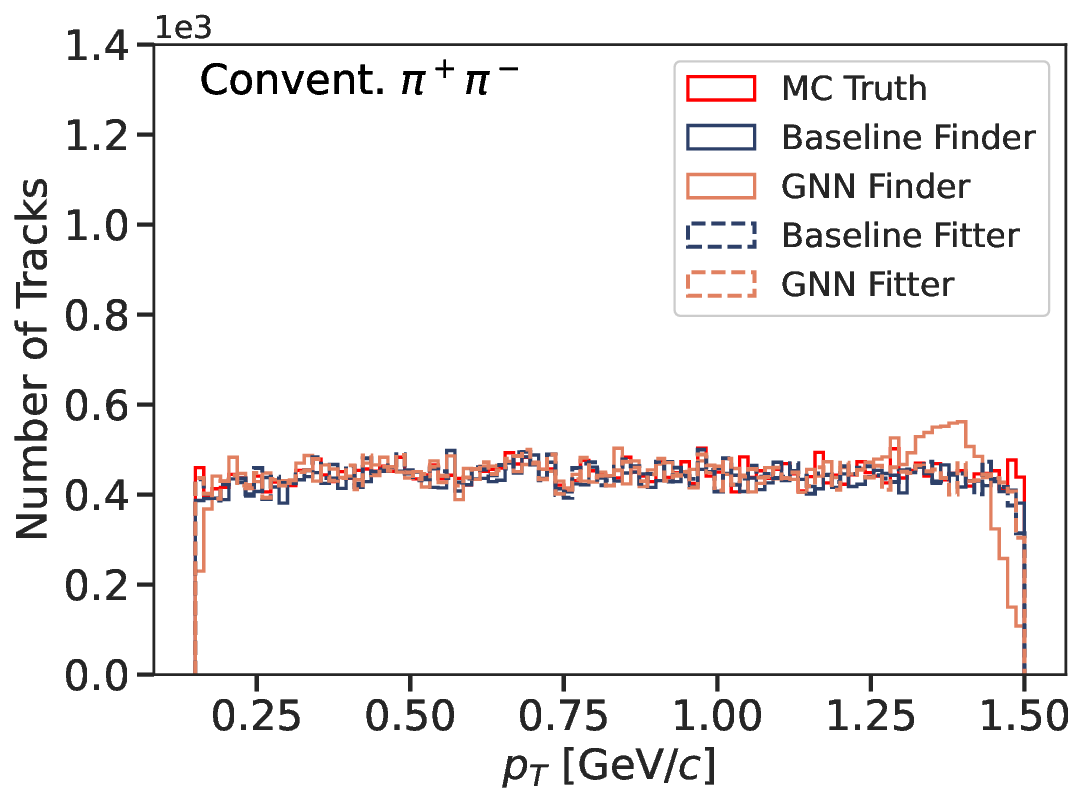}
  \end{minipage}
  \hfill
  \begin{minipage}[t]{0.3\textwidth}
    \centering
    \includegraphics[width=\linewidth]{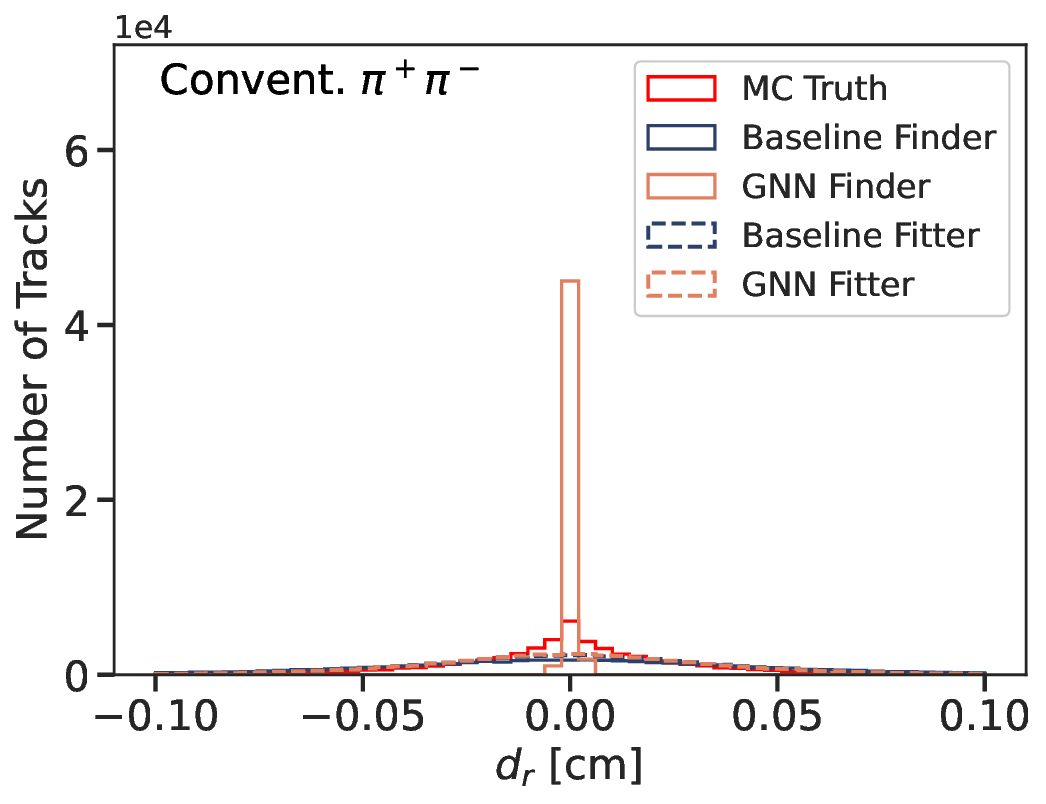}
  \end{minipage}
  \hfill
  \begin{minipage}[t]{0.31\textwidth}
    \centering
    \includegraphics[width=\linewidth]{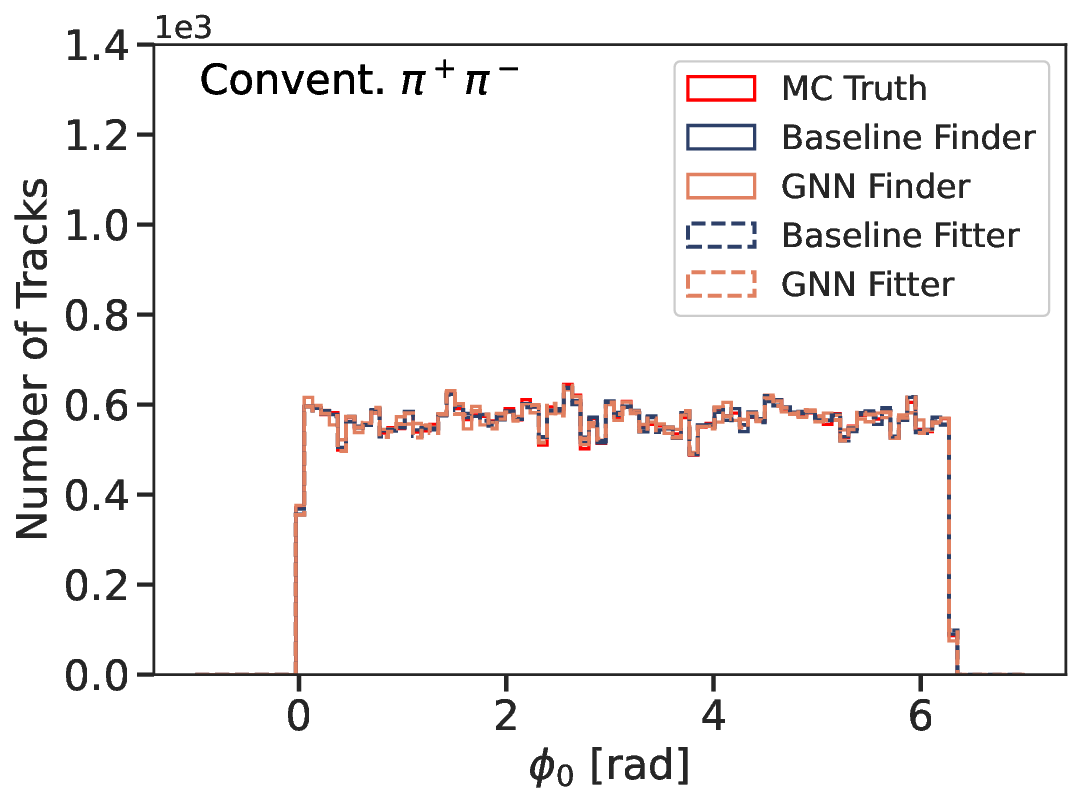}
  \end{minipage}
  
  \vspace{8pt}  % 带标注时可适当加大垂直间距
  
  % 第二行带标注
  \begin{minipage}[t]{0.3\textwidth}
    \centering
    \includegraphics[width=\linewidth]{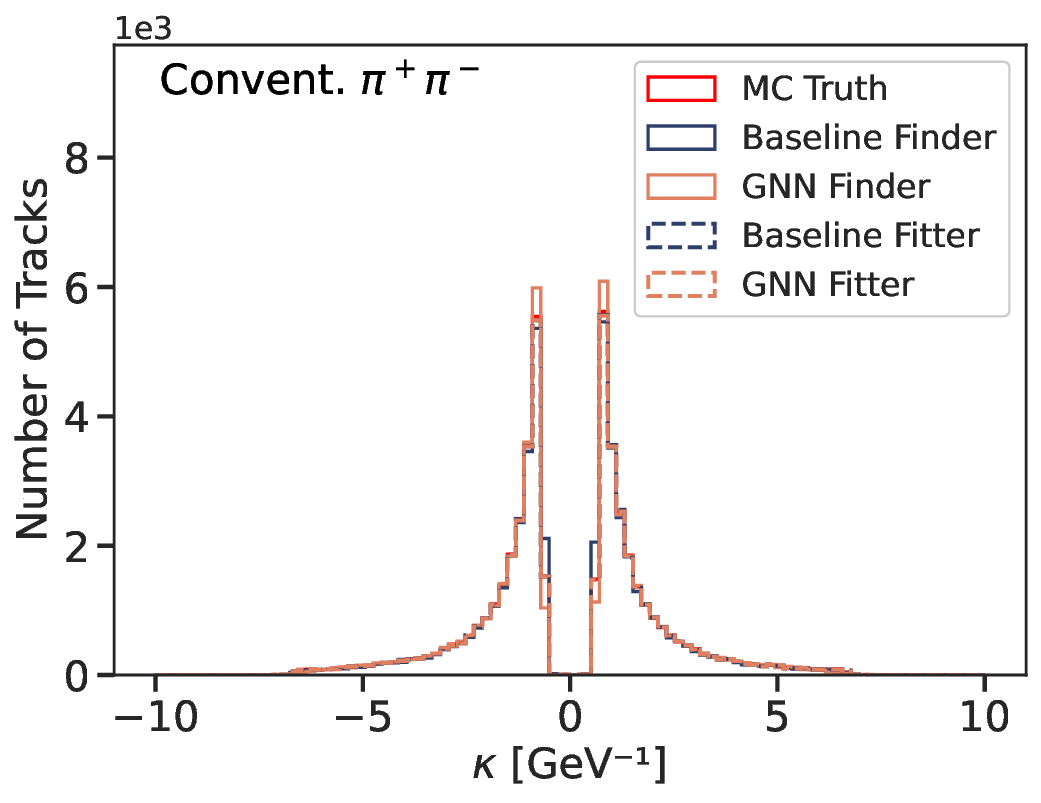}
  \end{minipage}
  \hfill
  \begin{minipage}[t]{0.31\textwidth}
    \centering
    \includegraphics[width=\linewidth]{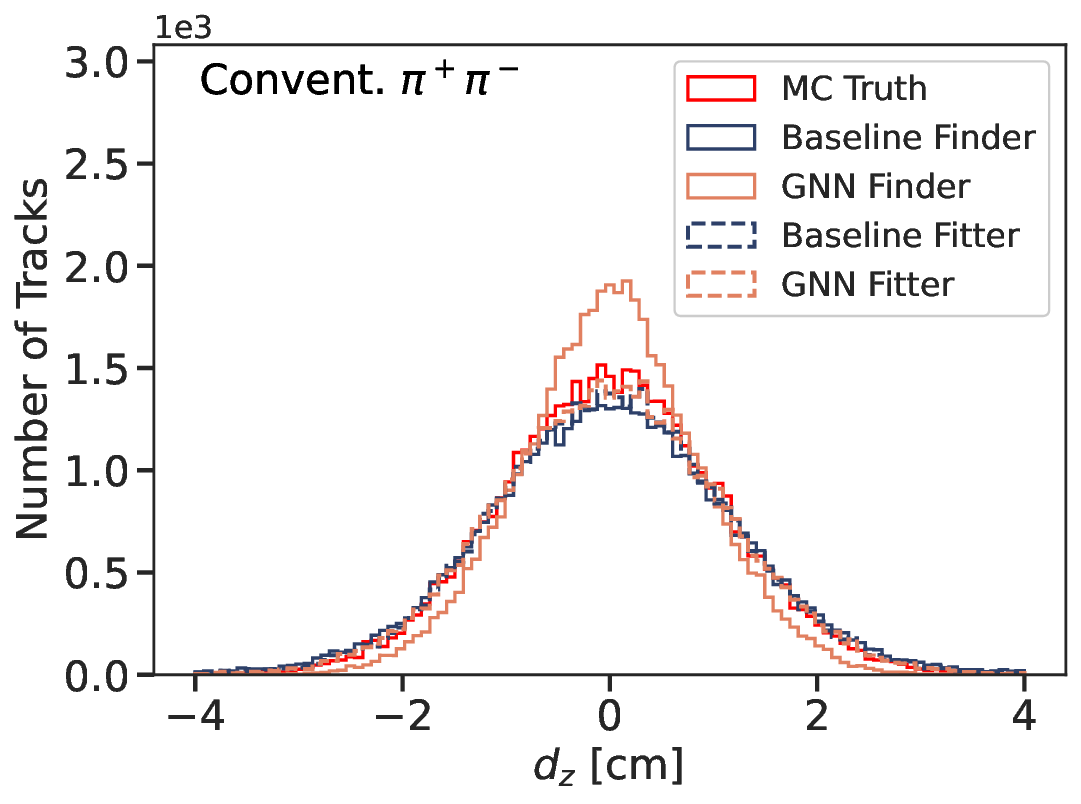}
  \end{minipage}
  \hfill
  \begin{minipage}[t]{0.31\textwidth}
    \centering
    \includegraphics[width=\linewidth]{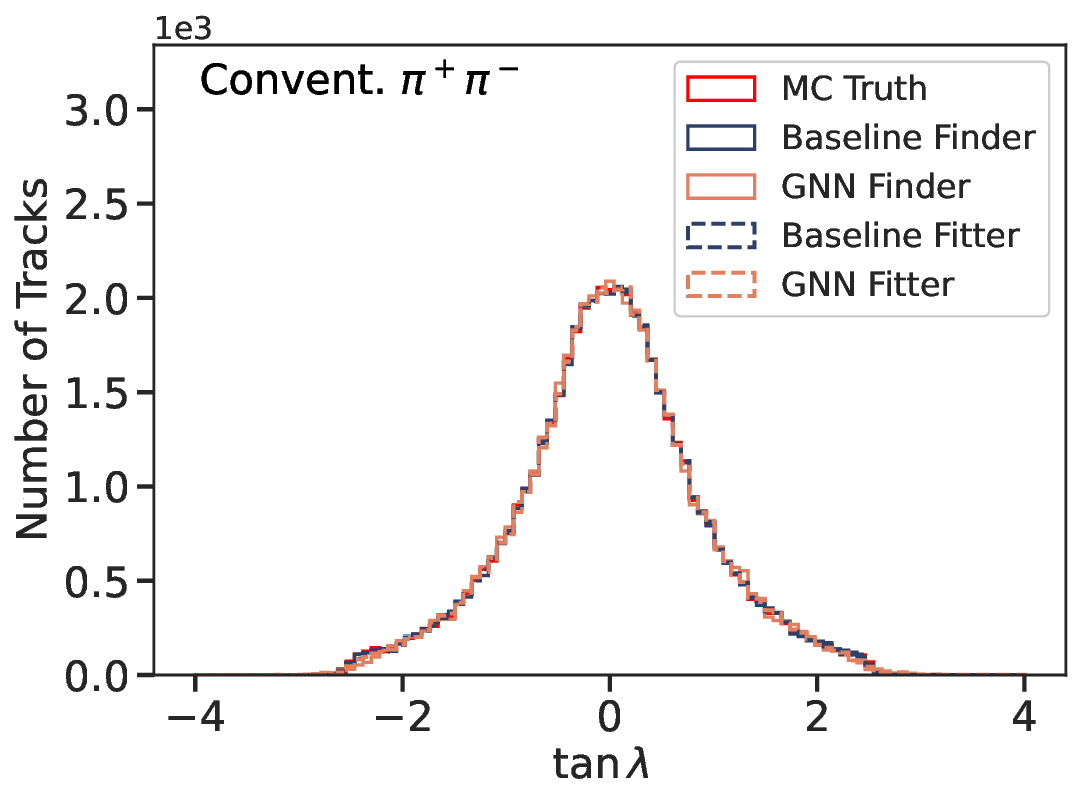}
  \end{minipage}

  \caption{Track parameters of \textit{conventional two-track} $\pi^+\pi^-$ events.}
  \label{fig:conven_track_parameters}
\end{figure}

\begin{figure}[!htbp]
  \centering
  % 第一行带标注
  \begin{minipage}[t]{0.31\textwidth}
    \centering
    \includegraphics[width=\linewidth]{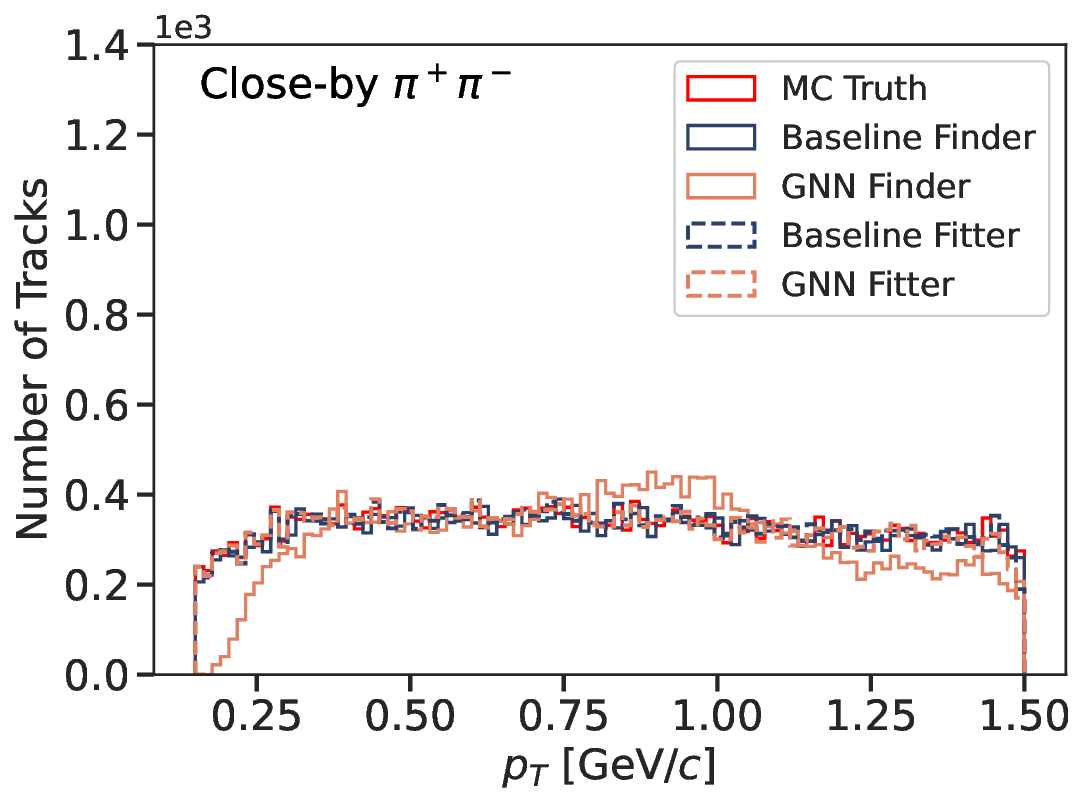}
  \end{minipage}
  \hfill
  \begin{minipage}[t]{0.3\textwidth}
    \centering
    \includegraphics[width=\linewidth]{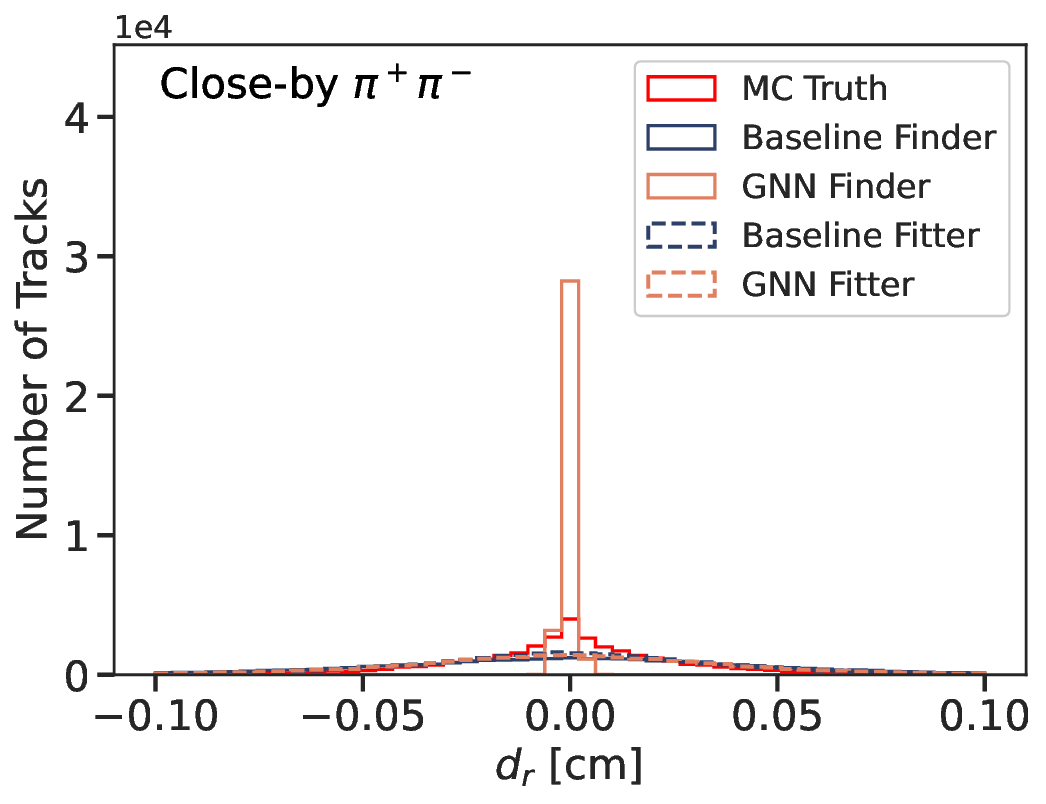}
  \end{minipage}
  \hfill
  \begin{minipage}[t]{0.31\textwidth}
    \centering
    \includegraphics[width=\linewidth]{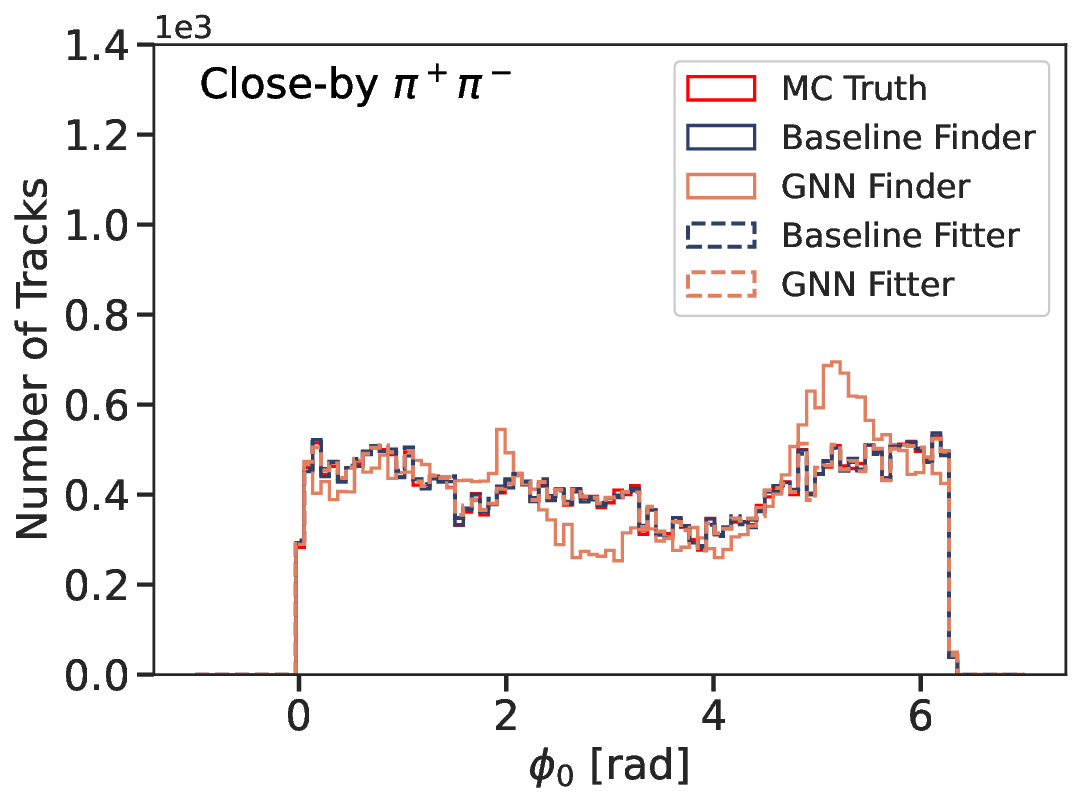}
  \end{minipage}
  
  \vspace{8pt}  % 带标注时可适当加大垂直间距
  
  % 第二行带标注
  \begin{minipage}[t]{0.31\textwidth}
    \centering
    \includegraphics[width=\linewidth]{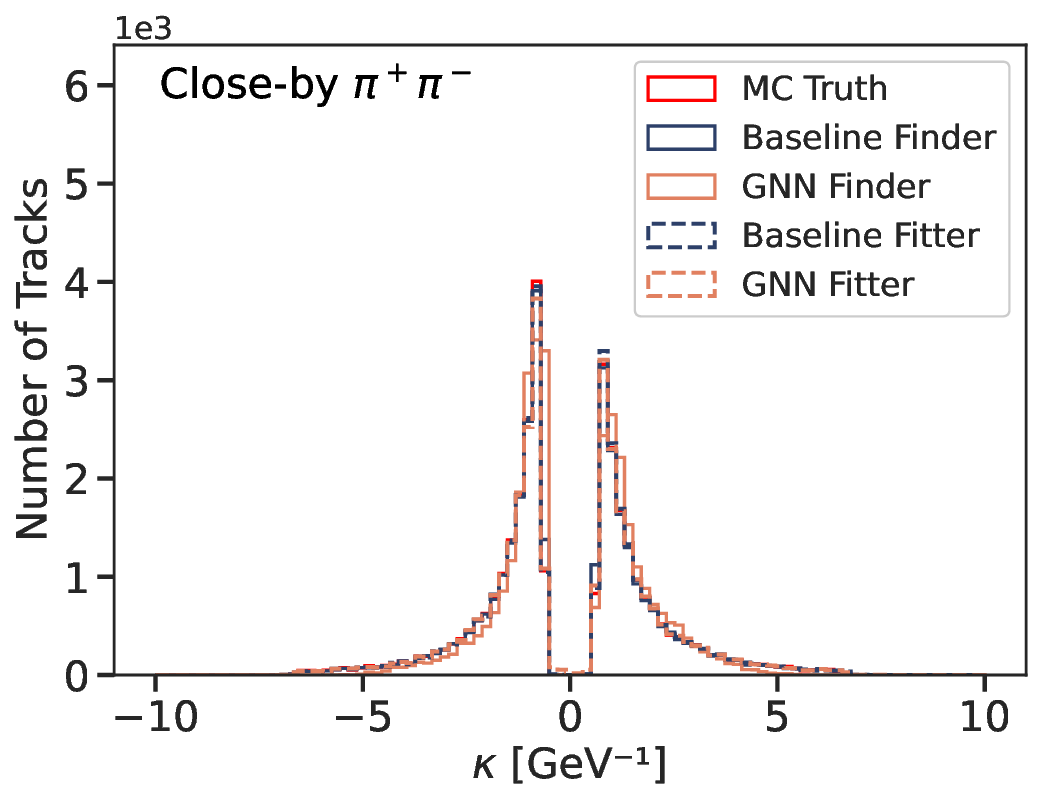}
  \end{minipage}
  \hfill
  \begin{minipage}[t]{0.33\textwidth}
    \centering
    \includegraphics[width=\linewidth]{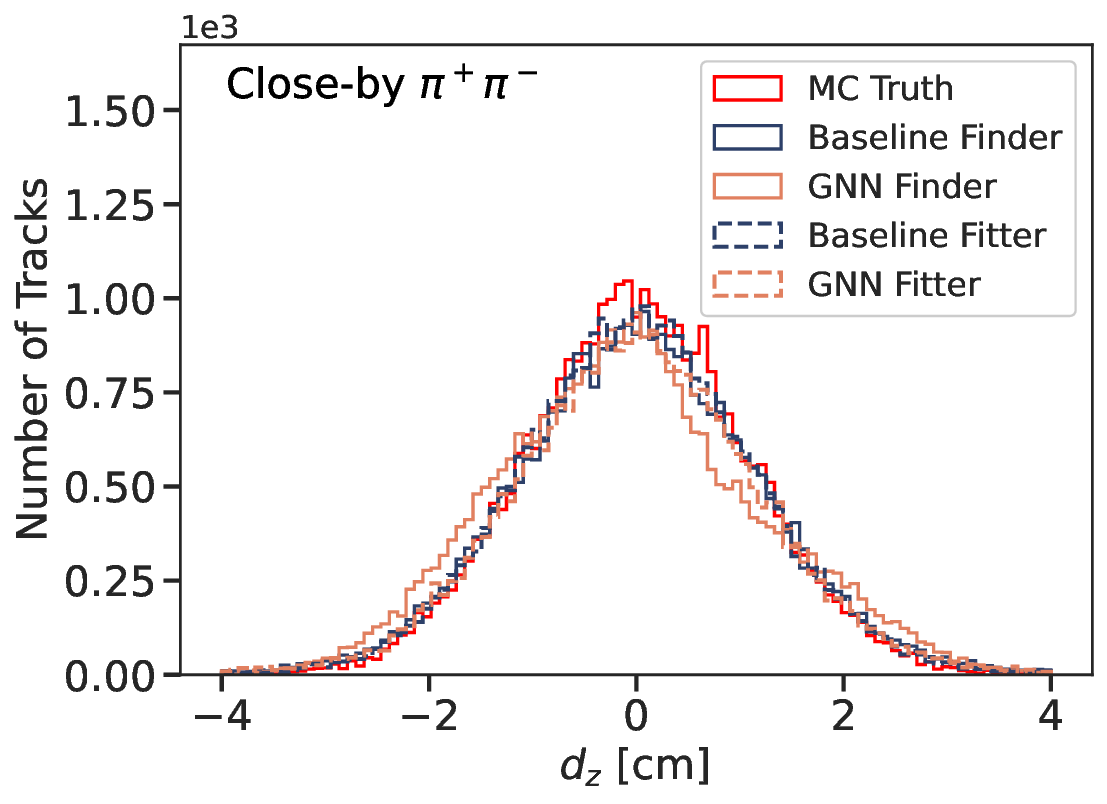}
  \end{minipage}
  \hfill
  \begin{minipage}[t]{0.31\textwidth}
    \centering
    \includegraphics[width=\linewidth]{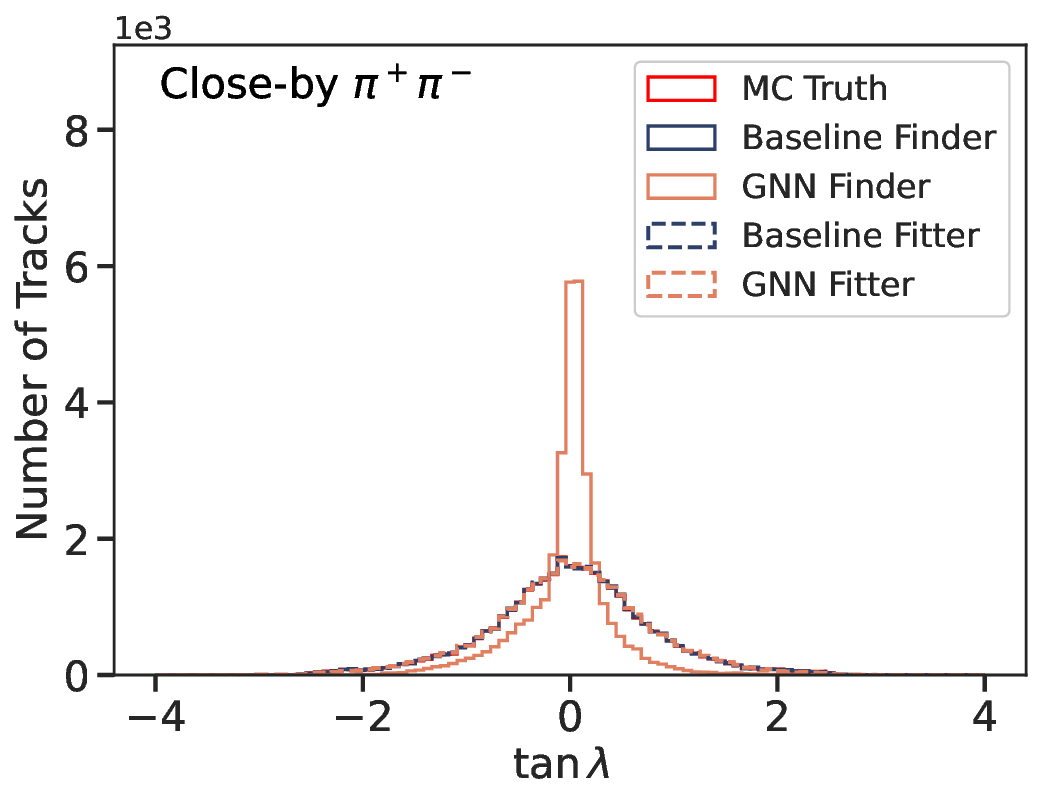}
  \end{minipage}

  \caption{Track parameters of \textit{close-by two-track} $\pi^+\pi^-$ events.}
  \label{fig:closeby_track_parameters}
\end{figure}

Figure~\ref{fig:single_track_parameters}, \ref{fig:conven_track_parameters} and \ref{fig:closeby_track_parameters} present the track parameters ($d_r$, $\phi_0$, $\kappa$, $d_z$ and $\tan\lambda$) of MC truth and tracks both found and fitted by the \baselinefinder and the \gnnfinder for \textit{single-track} $\pi^+$, \textit{conventional two-track} $\pi^+\pi^-$ and \textit{close-by two-track} $\pi^+\pi^-$ events, respectively. The track parameter distributions for other single-track particle species are analogous to those of $\pi^+$ and are thus omitted for brevity.

\newpage
\acknowledgments
\label{sec:acknowledgements}
We thank Mingrun Li for developing the pybes3 Python package, which supports the development of our evaluation metrics algorithm.

\textbf{Funding.} This work is supported by the National Natural Science Foundation of China (NSFC) under Contracts Nos. 12575207, 12175259, 12175124 and 12375197. Additional support was provided by the Strategic Priority Research Program of the Chinese Academy of Sciences under Grants Nos. XDA0480600 and XDA0480203.
% Zhang Yao RL 12575207, genfit 12175259
% Qin Xiaoshuai 12175124
% Zhang Jin 12375197
% Liu Beijiang, Li Ke XDA0480600
% Zhang Yi Fan XDA0480203

\textbf{Interests.} The authors declare that they have no conflict of interest.

% This is the most common positions for acknowledgments. A macro is
% available to maintain the same layout and spelling of the heading.

% \paragraph{Note added.} This is also a good position for notes added
% after the paper has been written.

% Bibliography
\bibliographystyle{JHEP}
\bibliography{biblio.bib}

\providecommand{\href}[2]{#2}\begingroup\raggedright\begin{thebibliography}{10}

\bibitem{CGEMupgrade}
I.~Balossino et~al., \emph{{The CGEM-IT: An Upgrade for the BESIII
  Experiment}}, \href{https://doi.org/10.3390/sym14050905}{\emph{Symmetry}
  {\bfseries 14} (2022) }.

\bibitem{BelleIIupgrade}
{\scshape BelleII} collaboration, \emph{{Snowmass Whitepaper: The Belle II
  Detector Upgrade Program}}, {\emph{arXiv} (2022) }
  [\href{https://arxiv.org/abs/arXiv:2203.11349}{{\ttfamily
  arXiv:2203.11349}}].

\bibitem{BESIIIlumi}
D.~Yin et~al., \emph{{Recent advances of experiment and simulation on
  luminosity performance at BEPCII}},
  \href{https://doi.org/https://doi.org/10.1016/j.nima.2025.170289}{\emph{Nuclear
  Instruments and Methods in Physics Research Section A} {\bfseries 1074}
  (2025) 170289}.

\bibitem{pattern}
Z.~Yao et~al., \emph{{Pattern-Matching Track Reconstruction for the BESIII Main
  Drift Chamber}}, {\emph{Chinese Physics C} {\bfseries 31} (2007) 570}.

\bibitem{fitting}
J.-K.~Wang et~al., \emph{{BESIII track fitting algorithm}},
  \href{https://doi.org/10.1088/1674-1137/33/10/010}{\emph{Chin. Phys. C}
  {\bfseries 33} (2009) 870}.

\bibitem{Correia:2025deq}
A.~Correia et~al., \emph{{Graph Neural Network-Based Pipeline for Track Finding
  in the Velo at LHCb}},  in \emph{{Connecting The Dots 2023}}, 6, 2025
  [\href{https://arxiv.org/abs/arXiv:2406.12869}{{\ttfamily
  arXiv:2406.12869}}].

\bibitem{Reuter:2024kja}
L.~Reuter et~al., \emph{{End-to-End Multi-track Reconstruction Using Graph
  Neural Networks at Belle~II}},
  \href{https://doi.org/10.1007/s41781-025-00135-6}{\emph{Comput. Softw. Big
  Sci.} {\bfseries 9} (2025) 6}.

\bibitem{Plini:2025hyu}
L.~Plini, G.~Tinti, T.~Spadaro and F.~Galasso, \emph{{Graph Neural Networks for
  particle tracking in NA62 Experiment}},
  \href{https://doi.org/10.1393/ncc/i2025-25149-3}{\emph{Nuovo Cim. C}
  {\bfseries 48} (2025) 149}.

\bibitem{GNN1}
L.~Heinrich et~al., \emph{{Combined track finding with GNN {\&} CKF}},  in
  \emph{{8th International Connecting the Dots Workshop (CTD 2023)}}, 1, 2024
  [\href{https://arxiv.org/abs/arXiv:2401.16016}{{\ttfamily
  arXiv:2401.16016}}].

\bibitem{GNN2}
J.~Duarte and J.-R.~Vlimant, \emph{{Graph Neural Networks for Particle Tracking
  and Reconstruction}},
  \href{https://arxiv.org/abs/[arXiv:2012.01249]}{{\ttfamily
  [arXiv:2012.01249]}}.

\bibitem{GNN3}
C.~Biscarat et~al., \emph{{Towards a realistic track reconstruction algorithm
  based on graph neural networks for the HL-LHC}},
  \href{https://doi.org/10.1051/epjconf/202125103047}{\emph{EPJ Web Conf.}
  {\bfseries 251} (2021) 03047}
  [\href{https://arxiv.org/abs/arXiv:2103.00916}{{\ttfamily
  arXiv:2103.00916}}].

\bibitem{trackML}
{Kaggle, CERN}, ``{TrackML Particle Tracking Challenge}.''
  \url{https://www.kaggle.com/competitions/trackml-particle-identification},
  2018.

\bibitem{TrackML1}
S.~Van~Stroud et~al., \emph{{Transformers for Charged Particle Track
  Reconstruction in High-Energy Physics}},
  \href{https://doi.org/10.1103/md46-yqgd}{\emph{Phys. Rev. X} {\bfseries 15}
  (2025) 041046} [\href{https://arxiv.org/abs/arXiv:2411.07149}{{\ttfamily
  arXiv:2411.07149}}].

\bibitem{TrackML2}
D.I.~Rusov et~al., \emph{{Deep Learning Methods in High Luminosity Track
  Reconstruction Scenario: Applying TrackNET to TrackML Challenge}},
  \href{https://doi.org/10.1134/S1063779625700935}{\emph{Phys. Part. Nucl.}
  {\bfseries 56} (2025) 1599}.

\bibitem{LHC}
O.~Br{\"u}ning and L.~Rossi, \emph{{The High Luminosity Large Hadron Collider
  -- HL-LHC}},  in \emph{Advanced Series on Directions in High Energy Physics},
  vol.~31, pp.~1--53, World Scientific (2024),
  \href{https://doi.org/10.1142/13487}{DOI}.

\bibitem{ODD1}
C.~Allaire et~al., ``{OpenDataDetector}.''
  \url{https://zenodo.org/records/6445359}, 4, 2022.
\newblock 10.5281/zenodo.6445359.

\bibitem{ColliderML}
D.~Elitez et~al., \emph{{ColliderML: The First Release of an OpenDataDetector
  High-Luminosity Physics Benchmark Dataset}},  2025.
\newblock 10.48550/arXiv.2512.15230.

\bibitem{BESIII}
{\scshape BESIII} collaboration, \emph{{Design and Construction of the BESIII
  Detector}}, \href{https://doi.org/10.1016/j.nima.2009.12.050}{\emph{Nucl.
  Instrum. Meth. A} {\bfseries 614} (2010) 345}.

\bibitem{BelleII}
I.H.~de~la Cruz, \emph{{The Belle II experiment: fundamental physics at the
  flavor frontier}},
  \href{https://doi.org/10.1088/1742-6596/761/1/012017}{\emph{Journal of
  Physics: Conference Series} {\bfseries 761} (2016) 012017}.

\bibitem{CEPC}
M.-Y.~Liu et~al., \emph{{Simulation and reconstruction of particle trajectories
  in the CEPC drift chamber}},
  \href{https://doi.org/10.1007/s41365-024-01497-z}{\emph{Nuclear Science and
  Techniques} {\bfseries 35} (2024) 128}.

\bibitem{STCF}
{\scshape STCF} collaboration, \emph{{STCF conceptual design report (Volume 1):
  Physics \& detector}},
  \href{https://doi.org/10.1007/s11467-023-1333-z}{\emph{Frontiers of Physics}
  {\bfseries 19} (2023) }.

\bibitem{COMET}
{\scshape COMET} collaboration, \emph{{Design and construction of the
  cylindrical drift chamber for the COMET Phase-I experiment}},
  \href{https://doi.org/https://doi.org/10.1016/j.nima.2024.169926}{\emph{Nuclear
  Instruments and Methods in Physics Research Section A} {\bfseries 1069}
  (2024) 169926}.

\bibitem{MEGII}
{\scshape MEGII} collaboration, \emph{{Towards a New
  \(\mu\)\(\rightarrow\)e\(\gamma\) Search with the MEG II Experiment: From
  Design to Commissioning}},
  \href{https://doi.org/10.3390/universe7120466}{\emph{Universe} {\bfseries 7}
  (2021) }.

\bibitem{FCC}
{\scshape FCC} collaboration, \emph{{Future Circular Collider Feasibility Study
  Report Volume 1: Physics and Experiments}},  2025.
\newblock 10.17181/CERN.9DKX.TDH9.

\bibitem{MDC}
{\scshape BESIII} collaboration, \emph{{Preliminary Design Report}: {The}
  {BESIII} {Detector}},  Tech. Rep. Institute of High Energy Physics, Chinese
  Academy of Sciences (2004).

\bibitem{BEPCII}
C.~Yu et~al., \emph{{BEPCII Performance and Beam Dynamics Studies on
  Luminosity}},  in \emph{7th International Particle Accelerator Conference},
  p.~TUYA01, 2016, \href{https://doi.org/10.18429/JACoW-IPAC2016-TUYA01}{DOI}.

\bibitem{BESIIIdata}
I.~Garzia, \emph{{Highlights from the BESIII experiment}},  in \emph{{QCD@Work
  2024: International Workshop on Quantum Chromodynamics - Theory and
  Experiment}}, vol.~314 of \emph{EPJ Web of Conferences}, p.~00008, 2024,
  \href{https://doi.org/10.1051/epjconf/202431400008}{DOI}.

\bibitem{Simulation}
Z.~Deng et~al., \emph{{BESIII simulation software}},
  \href{https://doi.org/10.22323/1.050.0043}{\emph{PoS} {\bfseries ACAT} (2007)
  043}.

\bibitem{BOSS}
J.~Zou, W.~Li, Q.~Ma et~al., \emph{Offline data processing system of the
  {BESIII} experiment},
  \href{https://doi.org/10.1140/epjc/s10052-024-13241-3}{\emph{Eur. Phys. J. C}
  {\bfseries 84} (2024) 937}.

\bibitem{HEPAI}
{Institute of High Energy Physics, Chinese Academy of Sciences (CAS)}, ``{High
  Energy Physics AI Platform}.'' \url{https://ai.ihep.ac.cn/}.

\bibitem{TSF}
Q.-G.~LIU et~al., \emph{{Track reconstruction using the TSF method for the
  BESIII main drift chamber}},
  \href{https://doi.org/10.1088/1674-1137/32/7/011}{\emph{Chinese Physics C}
  {\bfseries 32} (2008) 565}.

\bibitem{curlfinder}
L.-K.~Jia et~al., \emph{{{Study of low momentum track reconstruction for the
  {BESIII} main drift chamber}}},
  \href{https://doi.org/10.1088/1674-1137/34/12/014}{\emph{Chin. Phys. C}
  {\bfseries 34} (2010) 1866}.

\bibitem{hough}
J.~Zhang et~al., \emph{{Low transverse momentum track reconstruction based on
  the Hough transform for the BESIII drift chamber}},
  \href{https://doi.org/10.1007/s41605-018-0052-4}{\emph{Radiat. Detect.
  Technol. Methods} {\bfseries 2} (2018) 20}.

\bibitem{RK}
X.~Ai, H.M.~Gray, A.~Salzburger and N.~Styles, \emph{A non-linear kalman filter
  for track parameters estimation in high energy physics},
  \href{https://doi.org/10.1016/j.nima.2023.168041}{\emph{Nuclear Instruments
  and Methods in Physics Research Section A} {\bfseries 1049} (2023) 168041}.

\bibitem{genfit1}
C.~Höppner, S.~Neubert, B.~Ketzer and S.~Paul, \emph{A novel generic framework
  for track fitting in complex detector systems},
  \href{https://doi.org/https://doi.org/10.1016/j.nima.2010.03.136}{\emph{Nuclear
  Instruments and Methods in Physics Research Section A} {\bfseries 620} (2010)
  518}.

\bibitem{genfit2}
J.~Rauch and T.~Schl{\"u}ter, \emph{{GENFIT {\textemdash} a Generic
  Track-Fitting Toolkit}},
  \href{https://doi.org/10.1088/1742-6596/608/1/012042}{\emph{J. Phys. Conf.
  Ser.} {\bfseries 608} (2015) 012042}
  [\href{https://arxiv.org/abs/arXiv:1410.3698}{{\ttfamily arXiv:1410.3698}}].

\end{thebibliography}\endgroup

\end{document}